\documentclass[10pt,twocolumn,letterpaper]{article}

\usepackage{cvpr}              
\usepackage{multirow}
\usepackage{ulem}
\usepackage[export]{adjustbox}
\usepackage[table]{xcolor}
\usepackage{relsize}
\usepackage[most]{tcolorbox}

\usepackage[most]{tcolorbox}
\usepackage{svg}
\tcbset{
  myprompt/.style={
    breakable,
    colback=gray!5,        
    colframe=teal!60!black,
    boxrule=0.5pt,         
    arc=2mm,               
    left=4mm,right=4mm,    
    top=3mm,bottom=3mm,
    enhanced,
  }
}

\definecolor{cvprblue}{rgb}{0.21,0.49,0.74}
\usepackage[pagebackref,breaklinks,colorlinks,allcolors=cvprblue]{hyperref}

\makeatletter
\def\thanks#1{\protected@xdef\@thanks{\@thanks
        \protect\footnotetext{#1}}}
\makeatother

\def\paperID{6353} 
\def\confName{CVPR}
\def\confYear{2026}

\title{FFP-300K: Scaling First-Frame Propagation for Generalizable Video Editing}

\author{
Xijie Huang\textsuperscript{1*}, Chengming Xu\textsuperscript{2*}, Donghao Luo$^2$, Xiaobin Hu$^2$, Peng Tang$^2$, Xu Peng$^2$, Jiangning Zhang$^2$\\
Chengjie Wang$^2$, Yanwei Fu$^{1,3\dagger}$
\thanks{$*$: Equal contribution. $\dagger$: Corresponding author.}\\[2pt]
$^1$FDU,\quad
$^2$Tencent YouTu Lab,\quad
$^3$Shanghai Innovation Institute\\[2pt]
\href{ffp-300k.github.io}{\textit{ffp-300k.github.io}}\\
}
\begin{document}

\twocolumn[{%
    \renewcommand\twocolumn[1][]{#1}%
    \maketitle
    \begin{center}
	\centering
	\captionsetup{type=figure}
	\includegraphics[width=\textwidth]{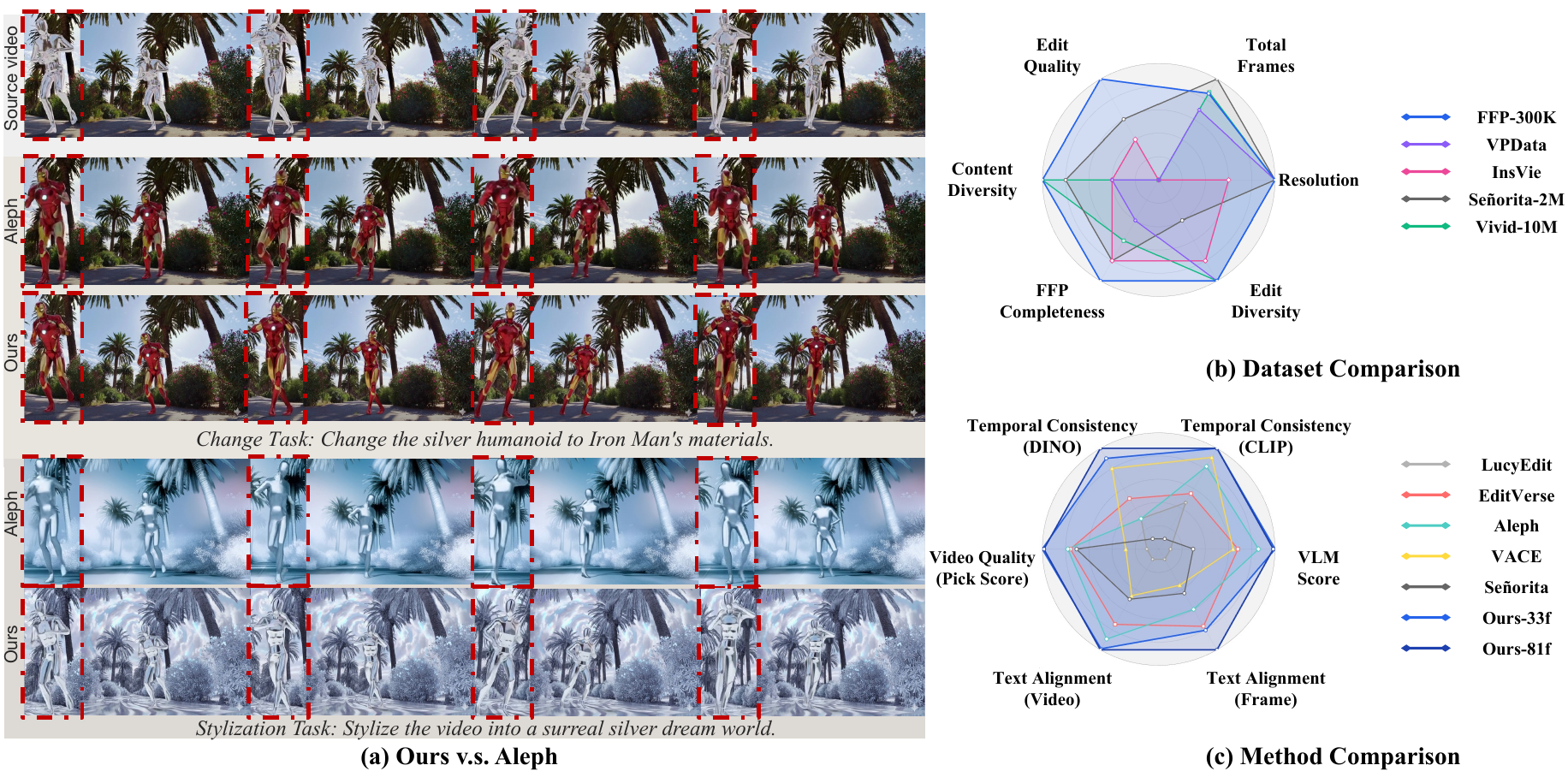}
	\vspace{-0.3in}
	\caption{
    (a) Results from our framework and Aleph~\cite{runway}, a commercial video editing model, with zoom-ins highlighting the main subject. 
    While generally capable, Aleph may \textbf{fail to follow the original motion} (top ``Change'' task) or \textbf{present limited visual quality} (bottom ``Stylization'' task), reflecting the capacity limits of current models. In comparison, based on a first frame edited by Qwen-Edit~\cite{qwenimage}, our framework achieves temporally consistent and visually realistic results on both tasks. For clarity, only Change and Stylization results are shown here; please see the supplementary material for more examples. 
    (b) Overall comparison between our proposed FFP-300K and previous video editing datasets. Each axis represents a key dataset aspect, including total frames for scale, resolution level, supported edit types, completeness of paired source–target data, content diversity across visual content and orientation types, and visual quality of generated target videos, providing an overall assessment of dataset scale, diversity, and consistency. Our FFP-300K is well suited for FFP-based video editing with higher-quality data. (c) Overall comparison between our framework and previous video editing methods, in which ours is generally better among all metrics.
    }
	\label{fig:teaser}
    \end{center}%
}]


\begin{abstract}
First-Frame Propagation (FFP) offers a promising paradigm for controllable video editing, but existing methods are hampered by a reliance on cumbersome run-time guidance. We identify the root cause of this limitation as the inadequacy of current training datasets, which are often too short, low-resolution, and lack the task diversity required to teach robust temporal priors. To address this foundational data gap, we first introduce FFP-300K, a new large-scale dataset comprising 300K high-fidelity video pairs at 720p resolution and 81 frames in length, constructed via a principled two-track pipeline for diverse local and global edits. Building on this dataset, we propose a novel framework designed for true guidance-free FFP that resolves the critical tension between maintaining first-frame appearance and preserving source video motion. Architecturally, we introduce Adaptive Spatio-Temporal RoPE (AST-RoPE), which dynamically remaps positional encodings to disentangle appearance and motion references. At the objective level, we employ a self-distillation strategy where an identity propagation task acts as a powerful regularizer, ensuring long-term temporal stability and preventing semantic drift. Comprehensive experiments on the EditVerseBench benchmark demonstrate that our method significantly outperforming existing academic and commercial models by receiving about 0.2 PickScore and 0.3 VLM score improvement against these competitors.

\end{abstract}

\section{Introduction}
\label{sec:intro}

High-fidelity video editing is a pivotal task with applications spanning professional film production, interactive entertainment, and the surge of user-generated content. An ideal model must provide users with precise control over edits while ensuring realism and temporal coherence. Current diffusion-based methods~\cite{dit, hunyuanvideo, wan2.1, cogvideox} largely follow two paradigms. 
The \textit{Instruction-based} approaches~\cite{Insv2v, LucyEdit2024}, while powerful for images, face compounded difficulty in the video domain. A model must simultaneously interpret a user's textual intent and apply it coherently across a temporal sequence, a dual challenge that often yields results that lag behind the fidelity of their image-based counterparts. In contrast, the \textit{First-Frame Propagation (FFP)} paradigm~\cite{i2vedit, stablev2v, genprop} offers a more pragmatic and powerful alternative by strategically decoupling the editing process. It allows users to leverage the sophisticated and mature ecosystem of image editing tools—from professional software to advanced generative models—to perfect a single frame with high precision. This approach alleviates the model's burden of semantic interpretation, transforming the complex task of text-to-video editing into a more constrained and well-defined problem: robust temporal propagation. However, this elegant promise of control is undermined by current models' reliance on cumbersome run-time guidance, such as per-video LoRA fine-tuning~\cite{ouyang2024i2vedit} or auxiliary inputs like depth maps~\cite{stablev2v}, which incur high computational costs and limit generalization.

This reliance on guidance is not a flaw in the FFP paradigm itself, but a symptom of inadequate training data. Lacking long, high-resolution, and diverse examples, models fail to learn robust temporal priors and are forced to use external guidance as a crutch. This data gap manifests in key limitations: (1) \textbf{Insufficient Length and Resolution:} Datasets like Señorita-2M~\cite{senorita} and InsViE~\cite{insvie} feature short, low-resolution clips, hindering the learning of long-range motion and fine details. (2) \textbf{Limited Task Diversity:} Many datasets focus on narrow tasks like inpainting (VPData~\cite{vpdata}) or fail to distinguish between local and global edits. (3) \textbf{Inconsistent Temporal Alignment:} Hybrid datasets like VIVID-10M~\cite{vivid-10m} mix images and videos, disrupting the learning of continuous motion priors.

To overcome these fundamental limitations, we introduce a synergistic solution comprising a new dataset and a novel framework. First, we present \textbf{FFP-300K}, a large-scale dataset engineered to directly address the aforementioned data challenges, which is constructed with a two-track synthesis pipeline. This pipeline leverages a motion-aware generative prior learned by VACE~\cite{vace} as its backbone to ensure temporal stability, employing mask-based manipulation for precise local edits and depth-guided conditioning for geometry-aware global stylization. This structured approach ensures task diversity and high fidelity. Benefited from the modularized pipeline, our dataset can be easily scaled up to provide sufficient generalization ability, which contains about \textbf{290,441} original/edited video pairs at \textbf{720p} resolution and a length of \textbf{81} frames, providing a rich and diverse foundation for training the next generation of video editing models.

Building upon FFP-300K, we then advance the FFP paradigm by proposing a new framework dubbed \textbf{FreeProp}, aiming to tackle the core challenge of \textit{balance between referencing the first frame for appearance and referencing the source video for motion} with two key contributions. For architectural level, we design an Adaptive Spatio-Temporal RoPE (AST-RoPE) that creates a content-aware geometry for the model. It learns from the source video to dynamically remap the spatio-temporal geometry, effectively disentangling the two references: it reduces the positional "distance" to the first frame to anchor appearance, while simultaneously rescaling the temporal axis to match the source video's motion. As for the objective level, we introduce a self-distillation strategy, in which the virtually created identity propagation task acts as a powerful regularizer, ensuring that the relational structure between the edited first frame and all subsequent frames follows a stable trajectory. This prevents semantic drift and ensures the edit's influence remains potent throughout the video.

We comprehensively evaluate our framework on the EditVerseBench~\cite{editverse} benchmark, demonstrating superior performance against recent models including both academic ones such as EditVerse~\cite{editverse} and commercial ones such as Aleph~\cite{runway} in both visual fidelity and temporal coherence. Our main contributions are:
\begin{itemize}
    \item We introduce \textbf{FFP-300K}, a large-scale dataset for FFP-based video editing, and the principled two-track generation pipeline used for its creation, addressing key limitations in prior data.
    \item We propose the novel Adaptive Spatio-Temporal RoPE (AST-RoPE) which disentangles appearance and motion.
    \item We introduce a powerful self-distillation strategy, which is crucial for maintaining the temporal stability and visual integrity required for guidance-free generation.
\end{itemize}

\section{Related Work}
\label{sec:relatedwork}
\noindent\textbf{Instruction-Based Video Editing Models.}
Instruction-based methods edit videos by interpreting natural language prompts. This paradigm is broadly divided into inversion-based and inversion-free approaches. Inversion-based models like VideoSwap~\cite{gu2024videoswap} and VideoDirector~\cite{wang2025videodirector} first map a source video into a latent noise space for editing. While this can yield precise results, the inversion process introduces significant computational overhead, limiting practical application. To circumvent this, inversion-free models are trained on large-scale datasets to generalize across diverse editing instructions. For instance, InsV2V~\cite{Insv2v} adapts image-to-image translation principles to video, while LucyEdit~\cite{LucyEdit2024} and EditVerse~\cite{editverse} introduce architectures to better integrate textual and visual conditioning. However, due to the intrinsic difficulty of this task, current instruction-based methods fall far behind their image counterparts.

\noindent\textbf{FFP-Based Video Editing Models.}
The FFP paradigm offers a more controllable alternative by decomposing video editing into two steps: user-driven first-frame modification and automated temporal propagation. Early methods like AnyV2V~\cite{anyv2v} and Videoshop~\cite{videoshop} demonstrated the potential of this approach but struggled with complex motion. Subsequent works sought to improve temporal coherence but introduced significant dependencies. For example, I2VEdit~\cite{i2vedit} requires costly per-video fine-tuning, rendering it unscalable. Others, like StableV2V~\cite{stablev2v} and GenProp~\cite{genprop}, rely on auxiliary guidance such as depth maps, optical flow, or predicted masks to preserve structure. Such reliance on external guidance complicates the pipeline and limits model generality due to dependence on auxiliary input quality. Our approach, by contrast, enables fully guidance-free propagation, i.e. conditioning solely on the source video and edited first frame to achieve temporally coherent and controllable results.

\noindent\textbf{Video Editing Datasets.}
The capabilities of video editing models are fundamentally shaped by the data they are trained on. Several large-scale datasets have been introduced to advance the field. Datasets like EffiVED~\cite{effived} and VPLM~\cite{raccoon} pioneered synthetic data generation for instruction-based tasks, while Señorita-2M~\cite{senorita}, VIVID-10M~\cite{vivid-10m}, VPData~\cite{vpdata}, and InsViE~\cite{insvie} significantly increased the scale and diversity of available data for object-level editing. Others such as IVEBench~\cite{chen2025ivebench} mainly focus on evaluation. However, existing datasets limit robust FFP model development with low-resolution, short clips and unclear distinctions between local and global edits. This forces models to rely on brittle, short-range priors, requiring the external guidance our method eliminates. Our FFP-300K dataset overcomes these issues with high-resolution (720p), long-form (81-frame) videos and separate tracks for local and global editing, establishing a standardized training set for generalizable FFP models.


\begin{figure*}[t]
  \centering
  \includegraphics[width=0.95\textwidth]{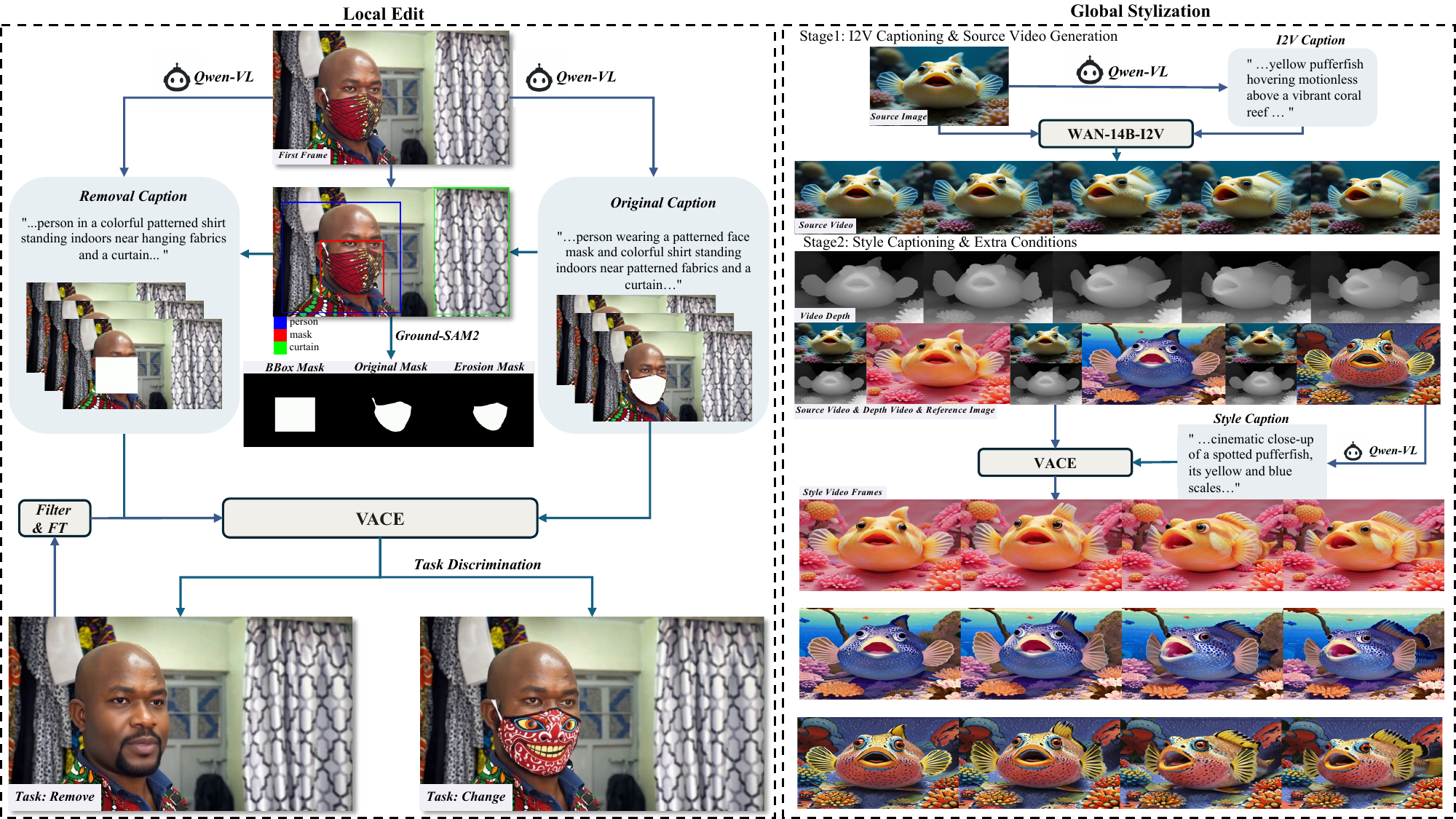}
  \vspace{-0.1in}
  \caption{
    \textbf{Overview of our Data Construction Pipeline}. Our pipeline has two parallel tracks. \textbf{Left: The local editing track} performs object Swap and Removal. For swapping, we use target objects and captions from the source video to generate edits with erosion masks, followed by a quality filtering step. For removal, captions are constructed and paired with bounding-box masks to generate the edited videos. Notably, filtered samples are used to refine our VACE~\cite{vace} model, which then regenerates the entire removal subset for higher quality (Sec.~\ref{sec:localediting}). \textbf{Right: The global stylization track} first generates source videos from images using Wan-I2V. It then combines these source videos, style reference images, and corresponding depth videos to produce high-fidelity stylized results (Sec.~\ref{sec:globalstylization}).
    }
  \vspace{-0.2in}
  \label{fig:framework}
\end{figure*}


\section{Scalable FFP Data Construction Pipeline}
\label{sec:data}

To address the need for a large-scale, high-fidelity dataset for FFP research, we construct \textbf{FFP-300K}. Our data generation framework is a \textbf{two-track modular pipeline} designed to produce semantically aligned video editing pairs at 720p resolution. Unlike unified pipelines, our framework operates via two independent and specialized branches to maximize quality for distinct editing categories:
\begin{enumerate}[label=(\arabic*), leftmargin=1.5em]
    \item \textbf{Local Editing:} Built upon the Koala-36M~\cite{koala}, this track focuses on fine-grained, object-level operations such as swapping and removal.
    \item \textbf{Global Stylization:} Derived from the Omni-Style~\cite{omnistyle}, this track emphasizes full-scene stylization.
\end{enumerate}
Each branch employs a tailored process of perception, captioning, and synthesis, culminating in a standardized dataset that supports both instruction-based and First-Frame Propagation (FFP) video editing frameworks.

\subsection{Local Editing} \label{sec:localediting}
The local editing branch generates precise object-level modifications. The process integrates large vision-language models (VLMs) for reasoning, advanced segmentation models for spatial localization, and a powerful video inpainting model for synthesis.

\noindent\textbf{Automated Editing Pipeline.}
For each source video from Koala-36M, we first use Qwen2.5-VL-72B-Instruct~\cite{qwen2} to analyze the first frame and identify primary editable objects. Subsequently, Grounded-SAM2~\cite{groundsam} performs instance segmentation to produce frame-wise mask videos, providing precise spatial constraints. These masks, along with task-specific captions, guide the video inpainting model VACE~\cite{vace} to synthesize the edit.
\begin{itemize}
    \item \textbf{For Swap tasks,} the original caption is used to guide VACE in replacing the masked object while preserving the background context.
    \item \textbf{For Removal tasks,} we prompt Qwen2.5-VL to generate a modified caption that explicitly describes the scene without the target object (e.g., "a street with a bench" instead of "a street with a person sitting on a bench"). This caption then guides VACE to remove the object and plausibly reconstruct the background.
\end{itemize}

\noindent\textbf{Refining Edits with Mask and Bounding Box Strategies.}
To optimize visual consistency, we discovered that the nature of the spatial conditioning is critical. We employ a \textbf{mask erosion strategy} to preserve only the boundary regions of the target mask, encouraging VACE to better leverage its internal priors for coherent inpainting. Furthermore, we experimented with two complementary conditioning schemes: providing VACE with only the eroded mask (\textit{without-bbox}) versus providing both the mask and the object's bounding box (\textit{with-bbox}). Our empirical analysis revealed a clear task-specific preference:
\begin{itemize}
    \item \textbf{Swap tasks} benefit from the \textit{without-bbox} approach, as the lack of a hard spatial constraint prevents artifacting and yields more semantically natural object integration.
    \item \textbf{Removal tasks} are more successful with the \textit{with-bbox} configuration, which provides a strong spatial prior that ensures complete object erasure and consistent background reconstruction.
\end{itemize}
This insight informs our quality control process, where we generate both high-quality variants.

\subsection{Global Stylization} \label{sec:globalstylization}
The global stylization branch transforms the entire visual appearance of a scene. Built upon the diverse Omni-Style dataset~\cite{omnistyle}, this track uses a two-stage process to ensure both semantic coherence and high stylistic fidelity.

\noindent\textbf{Stage 1: Source Video Generation.}
We first use Qwen2.5-VL to analyze each artistic image from Omni-Style and generate a cinematic video caption describing its scene, atmosphere, and tone. This caption is then used to prompt the Wan2.1-14B-I2V~\cite{wan2.1} to synthesize a source video, ensuring the generated motion and content are semantically aligned with the reference style image.

\noindent\textbf{Stage 2: Stylized Video Generation.}
Next, Qwen2.5-VL generates a detailed style caption by observing both the reference style image and the synthesized source video. This caption, which describes color palettes and textures, guides VACE in the stylization process. To preserve geometric structure, we provide VACE with depth maps extracted by Video Depth Anything~\cite{depthanything}. This combination of semantic guidance (style caption), structural guidance (depth), and appearance reference (style image) allows VACE to generate the final, temporally coherent stylized video.

\subsection{Quality Control and Curation}
A multi-stage filtering and verification process is applied to ensure the final dataset's quality and semantic integrity.

\noindent\textbf{Iterative Refinement for Removal Tasks.}
The removal subset underwent a particularly rigorous curation loop to maximize precision. First, Qwen2.5-VL automatically screens all generated videos for the removal task to filter out low-fidelity pairs, resulting in an initial set of nearly 40,000 candidates. This was followed by manual verification, yielding 14,389 high-quality samples. We then used this curated set to fine-tune the VACE model, significantly enhancing its removal capabilities. Finally, this improved VACE model was used to regenerate the entire removal subset, achieving cleaner background restoration.

\noindent\textbf{Final Verification and Statistics.}
All generated videos (swap, removal, and stylization) undergo a final semantic verification by Qwen2.5-VL to ensure a precise correspondence between the edit instruction and the visual transformation. After filtering and deduplication, our final FFP-300K dataset comprises 290,441 high-quality video pairs (source/edited video). This includes 143,913 stylization, 40,000 removal, and 106,528 swap/modification tasks. All videos are standardized to 720p resolution and 81 frames, making FFP-300K a robust and large-scale resource for advancing video editing research.

\section{Methodology}
\label{sec:method}

Our proposed FFP framework is designed to intrinsically handle temporal consistency, removing the need for explicit run-time conditions. The model is built upon a powerful conditional video model, which we adapt for the FFP task. Two core innovations, i.e. an adaptive positional encoding scheme and a self-distillation training objective, are introduced to resolve the core tension between appearance propagation and motion fidelity.

\begin{figure}[t]
  \centering
  \includegraphics[width=\linewidth]{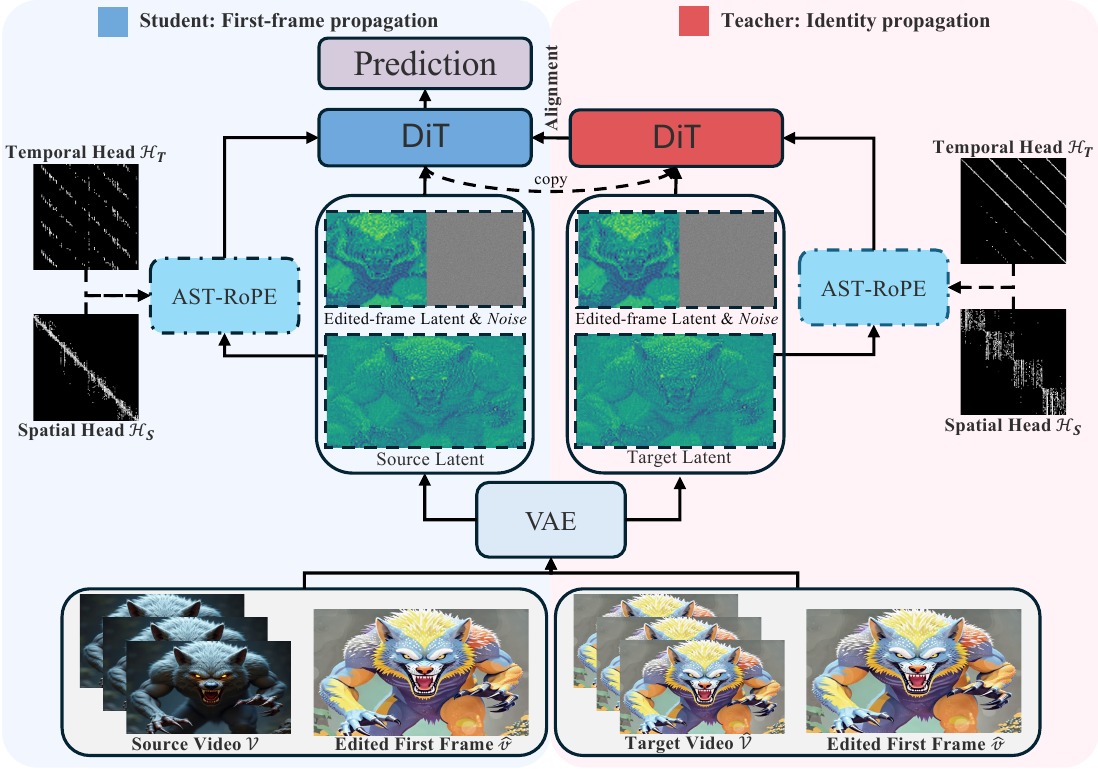}
  \vspace{-0.2in}
  \caption{
  \textbf{Overview of training paradigm.} Left: The source video and edited frame are encoded. The source latent informs our AST-RoPE module for adaptive spatio-temporal scaling. Right: The target video is processed identically to extract a latent DiT embedding, which is used to align the generation process.
  }
  \vspace{-0.2in}
  \label{fig:method}
\end{figure}

\subsection{Preliminary}

\noindent\textbf{Problem Formulation.} Our goal is the First-Frame Propagation (FFP) task: given a source video $\mathcal{V} \in \mathbb{R}^{F \times H \times W \times 3}$, where $F,H,W$ denotes number of frames, height and width respectively, and an edited first frame $\hat{v} \in \mathbb{R}^{H \times W \times 3}$, we aim to generate a target video $\hat{\mathcal{V}}$ that preserves the motion of $\mathcal{V}$ while propagating the edit from $\hat{v}$.

\noindent\textbf{Adapting Fun-Control for FFP.}
Our method is built upon Fun-Control, a powerful conditional video generation model derived from Wan 2.1~\cite{wan2.1}. Fun-Control is designed to aggregate conditioning information from both a video and a reference image, providing a strong prior for learning motion from the conditioning video while inheriting appearance from the conditioning image. This design is naturally suited for FFP, but was not originally designed for our specific task, presenting two key limitations:
\begin{enumerate}
    \item Its conditioning videos are low-level signals (e.g., depth maps), not the full RGB videos required in our case.
    \item Its reference images are often spatially unaligned, whereas in FFP, $\hat{v}$ is aligned in most regions.
\end{enumerate}
Therefore, task-specific adaptations are necessary. Formally, given $\mathcal{V}$ and $\hat{v}$, we first extract their corresponding VAE latents: $z_{src} \in \mathbb{R}^{F' \times H' \times W' \times C}$ and $\hat{z} \in \mathbb{R}^{H' \times W' \times C}$, where $C$ denotes feature channels. The first-frame latent $\hat{z}$ is then padded with zeros along the temporal dimension and concatenated with the noisy latent $z$, the source latent $z_{src}$, and a binary mask $M \in \mathbb{R}^{F' \times H' \times W'}$ (indicating the first frame) along the channel dimension. The resulting composite latent is fed into the DiT backbone for velocity prediction. By fine-tuning this model on our FFP-300K dataset using a flow matching objective, it effectively adapts its pre-trained motion prior to the specific requirements of FFP-based video editing.

\subsection{Adaptive Spatio-Temporal RoPE}
\label{sec:astrope}

The self-attention mechanism in a Diffusion Transformer (DiT) relies on Rotary Position Embeddings (RoPE)~\cite{su2024rope} to understand spatio-temporal relationships. However, standard RoPE imposes a static coordinate system that is ill-suited for FFP. Its uniform temporal progression is agnostic to the source video's intrinsic motion, and its fixed spatial distances hinder the propagation of the edited first frame, which must serve as a global content anchor.

To overcome this, we introduce \textbf{Adaptive Spatio-Temporal RoPE (AST-RoPE)}, a mechanism that endows the DiT with the ability to dynamically adapt its understanding of space and time based on the source video's content. Instead of a static grid, AST-RoPE learns to modulate the perceived positions of tokens, guiding self-attention to generate motion and appearance that is faithful to the source. This is achieved by predicting content-aware scaling coefficients that separately adjust the RoPE for specialized spatial and temporal self-attention heads.

\noindent\textbf{Source-Aware Scaling Coefficient Prediction.}
Inspired by the observation of head specialization in DiTs~\cite{xi2025sparse,ma2025follow}, we classify the attention heads in each layer into a static set of Spatial Heads ($\mathcal{H}_S$) and Temporal Heads ($\mathcal{H}_T$). For each video, a lightweight transformer module followed by a two-head MLP predicts a spatial scaling factor $\alpha_S$ and a temporal scaling factor $\alpha_T$ directly from the source latent $z_{src}$. This allows the model to infer high-level properties from the source video. For instance, predicting a smaller temporal scaling factor for a video with rapid motion.

We apply these coefficients distinctly to each head set. For spatial heads ($\mathcal{H}_S$), to enhance the first frame's influence, we use $\alpha_S$ to modulate its perceived positional distance. Specifically, the first item of temporal indice if offset from 0 to $\alpha_S\cdot F'$. By learning to predict $\alpha_S < 1$, we reduce the effective distance between the first frame and all other frames, especially the ending onees. This biases self-attention to assign higher scores between tokens in the edited first frame and those in subsequent frames, ensuring its content is robustly propagated.

For temporal heads ($\mathcal{H}_T$), we use $\alpha_T$ to rescale the temporal axis for all frames. The original temporal indices $[0, 1, \dots, F-1]$ are transformed to $[0, \alpha_T, \dots, \alpha_T(F-1)]$. This operation effectively stretches or compresses the temporal manifold. For a source video with rapid motion, the model can learn a smaller $\alpha_T$, reducing the perceived temporal distance between frames and encouraging the temporal heads to model more intense motion.

\subsection{Self-Distillation with Identity Propagation}
\label{sec:self_distill}


To enforce precise motion dynamics and first-frame reference, which standard flow matching fails to sufficiently constrain, we introduce a self-distillation paradigm. Our key insight is that the model's own internal processing of the source video provides the ideal alignment target. We implement this via a parallel \textbf{identity propagation} task, where the ``teacher'' task is to reconstruct the ground-truth target video $\hat{\mathcal{V}}$ from itself, i.e. conditioned on the $\hat{\mathcal{V}}$ and its first frame $\hat{v}$. This identity mapping forces its internal latents to perfectly encode the desired spatio-temporal dynamics. We then use distillation losses to align the standard ``student'' FFP task's representations with this idealized ``teacher'' representation, ensuring faithful motion preservation.

\noindent\textbf{Inter-Frame Relational Distillation.}
To ensure global motion patterns are preserved, we distill the frame-to-frame similarity structure, inspired by VideoREPA~\cite{zhang2025videorepa}. Given a latent representation $z^l \in \mathbb{R}^{F' \times H' \times W' \times C}$ from the $l$-th DiT block of the FFP task, and the corresponding latent $\hat{z}^l$ from the identity propagation task, we first downsample them spatially by a factor of $K_S$ to focus on motion over appearance. Let the resulting latents be $z^l_{ds}$ and $\hat{z}^l_{ds}$, with $N = (H'W')/K_S^2$ spatial tokens. Based on these two latents, the motion alignment can be calculate as:
\begin{align}
    G &= \textrm{Gram}(z^l_{ds}) \\
    \hat{G} &= \textrm{Gram}(\hat{z}^l_{ds}) \in\mathbb{R}^{F'\times N\times F'\times N} \\
    \mathcal{L}_{motion} &= \frac{1}{F'(F'-1)}\sum_{i,j=1}^{F'}\sum_{i\neq j}|G_{i,:,j.:}-\hat{G}_{i,:,j.:}|
\end{align}
where Gram denotes gram matrix along the channel dimension. $\mathcal{L}_{motion}$ minimizes the distance between the inter-frame relationships of the FFP latent and the identity propagation latent, which are indicative of motion, remain consistent with the source video's dynamics.

\begin{table*}[t]
\small
\centering
\resizebox{1\linewidth}{!}{
\begin{tabular}{c l c c c c c c c c}
\toprule
 & & & & \multicolumn{2}{c}{\textbf{Temporal Consistency}} & \multicolumn{2}{c}{\textbf{Text Alignment}} & \textbf{Video Quality} & \textbf{VLM Evaluation} \\
Type & Method & Resolution & Frames & CLIP $\uparrow$ & DINO $\uparrow$ & Frame $\uparrow$ & Video $\uparrow$ & Pick Score $\uparrow$ & VLM Score $\uparrow$ \\
\midrule
\multirow{2}{*}{\textbf{Training-free}}   &TokenFlow~\cite{qu2025tokenflow} &  640$\times$336 & 48 & 0.987 & 0.989 & 26.779 & 24.244 & 20.058 & 5.067 \\
 &   STDF~\cite{yatim2024stdf}     & 576$\times$320 & 24 & 0.965 & 0.964 & 26.422 & 23.768 & 19.817 & 4.911 \\
\midrule
\multirow{4}{*}{\textbf{Instruction-based}} &    InsV2V~\cite{Insv2v}  & 384$\times$384 & 32 & 0.972 & 0.969 & 25.923 & 23.092 & 19.611 & 5.252 \\
 & LucyEdit~\cite{LucyEdit2024}   & 832$\times$480 & 81 & 0.985 & 0.984 & 26.398 & 23.491 & 19.611 & 5.678 \\
 &  EditVerse~\cite{editverse}  & 624$\times$352 & 64 & 0.986 & 0.986 & 27.776 & 25.293 & 20.132 & 7.104 \\
&   Aleph~\cite{runway}   & 1280$\times$720 & 64 & 0.989 & 0.984 & 28.087 & 24.837 & 20.291 & 7.154 \\
\midrule
 \multirow{5}{*}{\textbf{FFP-based}} &   VACE~\cite{vace}    & 832$\times$480 & 61 & 0.990 & 0.989 & 27.169 & 24.188 & 20.095 & 6.072 \\
 &  Señorita~\cite{senorita}    & 864$\times$448 & 33 & 0.981 & 0.982 & 27.243 & 24.404 & 19.786 & 6.991 \\
 &  Señorita~\cite{senorita}$^*$    & 864$\times$448 & 33 & 0.989 & 0.987 & 27.754 & 24.657 & 19.913 & 7.341 \\
\rowcolor{gray!20} \cellcolor{white} & \textbf{Ours-33f}    & 1280$\times$720 & 33 & \textbf{0.991} & \uline{0.990} & \uline{28.293} & \uline{25.398} & \textbf{20.419} & \textbf{7.631} \\
\rowcolor{gray!20} \cellcolor{white} &  \textbf{Ours-81f}   & 1280$\times$720 & 81 & \textbf{0.991} & \textbf{0.991} & \textbf{28.316} & \textbf{25.925} & \uline{20.405} & \uline{7.600} \\
\bottomrule
\end{tabular}
}
\vspace{-0.1in}
\caption{\textbf{Quantitative comparison.} We compared three types of video editing methods on EditVerseBench. The best results are highlighted in \textbf{bold}, and the second-best results are \uline{underlined}. As shown, our \textbf{33f} and \textbf{81f} variants achieve the best performance across all automated evaluation metrics, establishing state-of-the-art results on EditVerseBench. Señorita$^*$ refers to using Qwen-Edit~\cite{qwenimage} to edit the first frame. 
}
\label{tab:nine_metrics_reordered}
\vspace{-0.2in}
\end{table*}

\noindent\textbf{First-Frame Consistency Loss.}
While motion alignment captures global structure, we need a focused mechanism to ensure the edit from the first frame propagates its influence consistently. We propose a novel loss based on Maximum Mean Discrepancy (MMD) to align the evolution of token-wise relationships with respect to the first frame.

For a given frame $i$, we compute the token-wise similarity matrix between the first frame and frame $i$: $S_i = z^l_1 (z^l_i)^T \in \mathbb{R}^{N \times N}$, where $z^l_1, z^l_i \in \mathbb{R}^{N \times C}$ are the (downsampled and reshaped) latents for the respective frames. Each of the $N$ rows of $S_i$ is a feature vector describing how a token in the first frame relates to all tokens in frame $i$. We treat this set of $N$ row vectors as an empirical distribution $P_i$ over an $N$-dimensional relation space.

We then use MMD with RBF kernel $k(\cdot, \cdot)$ to measure the divergence between the relational distribution of frame $i$ and that of the first frame (an identity relation), yielding a temporal drift score $d_i = \text{MMD}^2(P_1, P_i)$, along with the identity propagation counterpart $\hat{d}_i$, which are constrained to ensure similar evolution with each other:
\begin{equation}
    \mathcal{L}_{\text{MMD}} = \sum_{i=2}^{F} |d_i - \hat{d}_i|
\end{equation}
This loss regulates that the propagation of the first-frame edit follows a natural dynamic trajectory, as learned from the idealized identity task, preventing the edit's influence from fading or becoming distorted over time.

\noindent\textbf{Overall Training Objective.}
Our final training objective combines the standard flow matching loss $\mathcal{L}_{\text{FM}}$ with our two proposed distillation objectives:
\begin{equation}
    \mathcal{L} = \mathcal{L}_{\text{FM}} + \lambda_{\text{motion}}\mathcal{L}_{\text{motion}} + \lambda_{\text{MMD}}\mathcal{L}_{\text{MMD}},
\end{equation}
where $\lambda_{\text{motion}}$ and $\lambda_{\text{MMD}}$ are hyperparameters. Unlike methods that distill from external, generalist models~\cite{zhang2025videorepa}, our self-referential guidance is uniquely suited to FFP, as it distills from a teacher that has perfect knowledge of the source video's specific motion, ensuring edits are propagated without corrupting its essential temporal character.

\section{Experiments and Results}
\label{sec:exp}

\subsection{Experiment Setup}

\textbf{Implementation Details.} We finetune Fun-Control using LoRA~\cite{lora} for 2 epoches with rank is set to 128.
AdamW~\cite{loshchilov2017adamw} is utilized for training with a learning rate of $2 \times 10^{-4}$ and cosine decay. 
$\lambda_{\text{motion}}$ and $\lambda_{\text{MMD}}$ are set to 5 and 1.
For fair comparison with previous methods, for the main experiments we train two variants of our model with 81-frame videos and 33-frame videos. For ablation study, the one trained with 81-frame videos is engaged in.


\noindent\textbf{Benchmark and Metrics.} For evaluation, we adopt EditVerseBench~\cite{editverse}, a comprehensive benchmark for video editing that covers 20 diverse editing categories.
Since our method focuses on FFP-based video editing, we further filter the benchmark to 125 videos with stable temporal structures that can be evaluated under a propagation setting. Then Qwen-Edit~\cite{qwenimage} is leveraged to generate the edited first frame for these videos. We follow the six metrics defined in EditVerseBench: VLM editing quality, PickScore, Frame score, Video score, CLIP text–image alignment, and DINO-based temporal consistency.
To better assess long-sequence propagation, we extend the VLM evaluation from 2 frames to 10 sampled frames.
Different from the original setup that uses GPT-4o~\cite{hurst2024gpt4o} as the evaluation model, we replace it with Qwen2.5-VL-72B-Instruct~\cite{bai2025qwen2.5vl} to ensure full reproducibility and consistency across evaluation runs.

\noindent\textbf{Competitors.} We directly adopt the baseline models provided in EditVerseBench\cite{editverse}, including Token-Flow~\cite{qu2025tokenflow}, STDF~\cite{yatim2024stdf}, InsV2V~\cite{Insv2v}, Lucy-Edit~\cite{LucyEdit2024}, Señorita-2M~\cite{senorita}, and Aleph~\cite{runway}, to ensure a fair and consistent comparison.

\subsection{Quantitative Comparison}
In Tab.~\ref{tab:nine_metrics_reordered} we present the quantitative comparison between the competitors and two variants of our method. For fair comparison, we test Senorita with the same edited first frames as ours.
Our method, in both 33-frame (\textbf{Ours-33f}) and 81-frame (\textbf{Ours-81f}) configurations, consistently outperforms all competing approaches across the board. Specifically, Ours-81f achieves the highest scores in temporal consistency (0.991 CLIP score, 0.991 DINO score) and video-level text alignment (25.925), showcasing its exceptional ability to maintain coherence over longer sequences. Furthermore, Ours-33f obtains the top scores in perceptual quality (20.419 Pick Score) and semantic correctness (7.631 VLM Score), indicating superior alignment with user intent. Notably, our model surpasses not only other FFP-based methods like VACE~\cite{vace} but also strong instruction-based models, including the commercially used Aleph~\cite{runway}. This highlights the effectiveness of our approach in achieving a superior balance of temporal stability, edit fidelity, and overall visual quality.

\begin{figure*}[t]
  \centering
  \includegraphics[width=\textwidth]{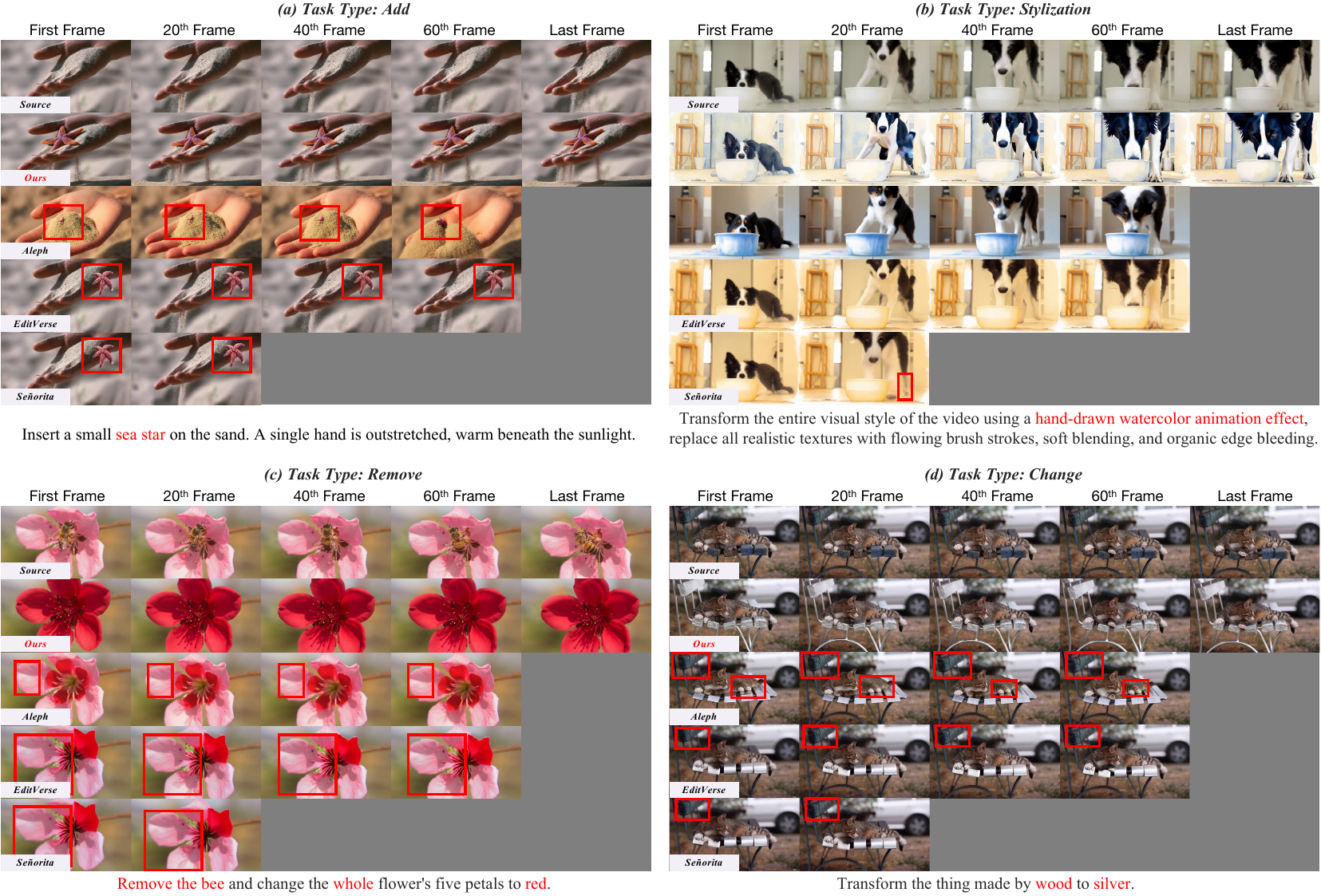}
  \vspace{-0.15in}
  \caption{
    \textbf{Qualitative comparison.} We Choose top three method in quantitative comparison to compare with our visual results across four representative video editing tasks. Red boxes highlight the unreasonable generated contents. The gray placeholder denotes these methods cannot generate such long videos. Our method generally enjoys better editing fidelity, temporal consistency and visual quality.
  }
  \vspace{-0.2in}
  \label{fig:qualitative}
\end{figure*}

\subsection{Qualitative Comparison}
We further provide qualitative results to visually demonstrate the advantages of our framework.
As illustrated in Fig.~\ref{fig:qualitative}, previous instruction-based methods such as Aleph and EditVerse mainly suffer from the problem of unsuitable edited first frame, such as the wrong position of starfish in Fig.~\ref{fig:qualitative}(a), and failure to preserve the content in the original videos. Moreover, it is noteworthy that videos generated by EditVerse also has the flickering problem, which cannot be fully presented with static frames but will be shown in the supplementary material. On the other hand, Senorita, as a FFP-based method, is limited with the video quality, showing mosaic in the bottom of each frame. In contrast, our method produces results with not only longer duration, but also significantly better general quality, accurately preserving object structure and scene layout while maintaining global temporal consistency. This qualitative superiority verifies that the combination of our proposed techniques for consistency modeling and curated high-fidelity dataset enables robust editing propagation across diverse and challenging real-world scenarios.

\subsection{User Study}

To further evaluate perceptual quality and editing accuracy, we conducted a user study where participants rated videos on a 1–5 scale based on: (1) \textbf{Editing Accuracy (EA)}: instruction adherence and semantic consistency, (2) \textbf{Motion Accuracy (MA)}: motion fidelity to the source video, and (3) \textbf{Video Quality (VQ)}: temporal smoothness and realism. With 15 participants each assessing 8 random videos from EditVerseBench, our method achieved the highest mean scores in all criteria (Tab.~\ref{tab:userstudy}), demonstrating user preference for its precise alignment and stable dynamics, consistent with quantitative results.

\begin{table}[htb]
\centering
\resizebox{0.65\linewidth}{!}{
\small
\begin{tabular}{lccc}
\toprule
\textbf{Methods} & \textbf{EA $\uparrow$} & \textbf{MA $\uparrow$} & \textbf{VQ $\uparrow$} \\
\midrule
EditVerse~\cite{editverse}          & 4.063 & 3.792 & 3.354 \\
Señorita-2M~\cite{senorita}     & 3.563 & 3.208 & 2.354 \\
Aleph~\cite{runway}    & 3.412 & 3.271 & 3.459 \\
\textbf{Ours}   & \textbf{4.250} & \textbf{4.333} & \textbf{4.146} \\
\bottomrule
\end{tabular}
}
\vspace{-0.1in}
\caption{User study preference regarding editing accuracy (EA), motion accuracy (MA) and video quality (VQ). Our method is consistently preferred.}
\label{tab:userstudy}
\vspace{-0.2in}
\end{table}

\begin{table}[htb]
\small
\centering
\resizebox{1\linewidth}{!}{
\begin{tabular}{l c c c c c c c c}
\toprule
   & \multicolumn{2}{c}{\textbf{Temporal Consistency}} & \multicolumn{2}{c}{\textbf{Text Alignment}} & \textbf{Video Quality} & \textbf{VLM Evaluation} \\
 Method & CLIP $\uparrow$ & DINO $\uparrow$ & Frame $\uparrow$ & Video $\uparrow$ & Pick Score $\uparrow$ & VLM Score $\uparrow$ \\
\midrule
Baseline  & 0.986 & 0.984 & 27.420 & 24.960 & 20.010 & 7.210 \\
+AST-RoPE               & 0.989 & 0.988 & 28.178 & 25.817 & 20.354 & 7.542 \\
Full     & \textbf{0.991} & \textbf{0.991} & \textbf{28.316} & \textbf{25.925} & \textbf{20.405} & \textbf{7.600} \\
\bottomrule
\end{tabular}
}
\vspace{-0.1in}
\caption{Quantitative results for ablation variants of our model.}
\label{tab:ablation}
\vspace{-0.2in}
\end{table}

\subsection{Ablation Study}
To validate the efficacy of each component in our framework, we conduct ablation study on three variants trained with 81-frame videos: (1) \textbf{Baseline}: the original Wan-Fun model fine-tuned on our dataset without any modification, (2) \textbf{+AST-RoPE}: applying our spatial–temporal RoPE adaptation to enhance attention modules, (3) \textbf{Full}: integrating both RoPE adaptation and our proposed self-distillation strategy.
Results are summarized in Tab.~\ref{tab:ablation}. Thanks to our proposed dataset, the baseline model can already achieve strong performance. Based on that, the RoPE adaptation and self-distillation can further enhance the quality in terms of both visual quality and text alignment, indicating the effectiveness of the proposed method.


\section{Conclusion}
\label{sec:conclsion}

We addressed a core limitation in First-Frame Propagation (FFP) video editing: a foundational data gap that necessitates complicated run-time guidance which results in limited generalization ability, for which our solution covers two main aspects. First, we introduce \textbf{FFP-300K}, a large-scale dataset with high-quality and diverse videos. Second, our model leverages a novel Adaptive Spatio-Temporal RoPE (AST-RoPE) and self-distillation to strengthen the first-frame reference and source motion preservation. This dual approach achieves state-of-the-art fidelity and temporal coherence. By tackling both data and model, we make high-fidelity, controllable video editing a practical reality.

{
    \small
    \bibliographystyle{ieeenat_fullname}
    \bibliography{main}

\begin{thebibliography}{41}
\providecommand{\natexlab}[1]{#1}
\providecommand{\url}[1]{\texttt{#1}}
\expandafter\ifx\csname urlstyle\endcsname\relax
  \providecommand{\doi}[1]{doi: #1}\else
  \providecommand{\doi}{doi: \begingroup \urlstyle{rm}\Url}\fi

\bibitem[run()]{runway}
Introducing runway aleph.

\bibitem[Bai et~al.(2025)Bai, Chen, Liu, Wang, Ge, Song, Dang, Wang, Wang, Tang, et~al.]{bai2025qwen2.5vl}
Shuai Bai, Keqin Chen, Xuejing Liu, Jialin Wang, Wenbin Ge, Sibo Song, Kai Dang, Peng Wang, Shijie Wang, Jun Tang, et~al.
\newblock Qwen2. 5-vl technical report.
\newblock \emph{arXiv preprint arXiv:2502.13923}, 2025.

\bibitem[Bian et~al.(2025)Bian, Zhang, Ju, Cao, Xie, Shan, and Xu]{vpdata}
Yuxuan Bian, Zhaoyang Zhang, Xuan Ju, Mingdeng Cao, Liangbin Xie, Ying Shan, and Qiang Xu.
\newblock Videopainter: Any-length video inpainting and editing with plug-and-play context control.
\newblock In \emph{Proceedings of the Special Interest Group on Computer Graphics and Interactive Techniques Conference Conference Papers}, pages 1--12, 2025.

\bibitem[Chen et~al.(2025)Chen, Zhang, Hu, Zeng, Xue, He, Wang, Liu, Hu, and Yan]{chen2025ivebench}
Yinan Chen, Jiangning Zhang, Teng Hu, Yuxiang Zeng, Zhucun Xue, Qingdong He, Chengjie Wang, Yong Liu, Xiaobin Hu, and Shuicheng Yan.
\newblock Ivebench: Modern benchmark suite for instruction-guided video editing assessment.
\newblock \emph{arXiv preprint arXiv:2510.11647}, 2025.

\bibitem[Cheng et~al.(2024)Cheng, Xiao, and He]{Insv2v}
Jiaxin Cheng, Tianjun Xiao, and Tong He.
\newblock Consistent video-to-video transfer using synthetic dataset.
\newblock In \emph{The Twelfth International Conference on Learning Representations}, 2024.

\bibitem[Fan et~al.(2024)Fan, Bhattad, and Krishna]{videoshop}
Xiang Fan, Anand Bhattad, and Ranjay Krishna.
\newblock Videoshop: Localized semantic video editing with noise-extrapolated diffusion inversion.
\newblock In \emph{European Conference on Computer Vision}, pages 232--250. Springer, 2024.

\bibitem[Gu et~al.(2024)Gu, Zhou, Wu, Yu, Liu, Zhao, Wu, Zhang, Shou, and Tang]{gu2024videoswap}
Yuchao Gu, Yipin Zhou, Bichen Wu, Licheng Yu, Jia-Wei Liu, Rui Zhao, Jay~Zhangjie Wu, David~Junhao Zhang, Mike~Zheng Shou, and Kevin Tang.
\newblock Videoswap: Customized video subject swapping with interactive semantic point correspondence.
\newblock In \emph{Proceedings of the IEEE/CVF Conference on Computer Vision and Pattern Recognition}, pages 7621--7630, 2024.

\bibitem[Hu et~al.(2022)Hu, Shen, Wallis, Allen-Zhu, Li, Wang, Wang, Chen, et~al.]{lora}
Edward~J Hu, Yelong Shen, Phillip Wallis, Zeyuan Allen-Zhu, Yuanzhi Li, Shean Wang, Lu Wang, Weizhu Chen, et~al.
\newblock Lora: Low-rank adaptation of large language models.
\newblock \emph{ICLR}, 1\penalty0 (2):\penalty0 3, 2022.

\bibitem[Hu et~al.(2025)Hu, Zhong, Wang, Jiang, Tian, Yang, Wan, and Zhang]{vivid-10m}
Jiahao Hu, Tianxiong Zhong, Xuebo Wang, Boyuan Jiang, Xingye Tian, Fei Yang, Pengfei Wan, and Di Zhang.
\newblock Vivid-10m: A dataset and baseline for versatile and interactive video local editing, 2025.

\bibitem[Hurst et~al.(2024)Hurst, Lerer, Goucher, Perelman, Ramesh, Clark, Ostrow, Welihinda, Hayes, Radford, et~al.]{hurst2024gpt4o}
Aaron Hurst, Adam Lerer, Adam~P Goucher, Adam Perelman, Aditya Ramesh, Aidan Clark, AJ Ostrow, Akila Welihinda, Alan Hayes, Alec Radford, et~al.
\newblock Gpt-4o system card.
\newblock \emph{arXiv preprint arXiv:2410.21276}, 2024.

\bibitem[Jiang et~al.(2025)Jiang, Han, Mao, Zhang, Pan, and Liu]{vace}
Zeyinzi Jiang, Zhen Han, Chaojie Mao, Jingfeng Zhang, Yulin Pan, and Yu Liu.
\newblock Vace: All-in-one video creation and editing.
\newblock \emph{arXiv preprint arXiv:2503.07598}, 2025.

\bibitem[Ju et~al.(2025)Ju, Wang, Zhou, Zhang, Liu, Zhao, Zhang, Li, Cai, Liu, Pakhomov, Lin, Kim, and Xu]{editverse}
Xuan Ju, Tianyu Wang, Yuqian Zhou, He Zhang, Qing Liu, Nanxuan Zhao, Zhifei Zhang, Yijun Li, Yuanhao Cai, Shaoteng Liu, Daniil Pakhomov, Zhe Lin, Soo~Ye Kim, and Qiang Xu.
\newblock Editverse: Unifying image and video editing and generation with in-context learning, 2025.

\bibitem[Kong et~al.(2025)Kong, Tian, Zhang, Min, Dai, Zhou, Xiong, Li, Wu, Zhang, Wu, Lin, Yuan, Long, Wang, Wang, Li, Huang, Yang, Tan, Wang, Song, Bai, Wu, Xue, Wang, Wang, Liu, Li, Li, Wang, Yu, Deng, Li, Chen, Cui, Peng, Yu, He, Xu, Zhou, Xu, Tao, Lu, Liu, Zhou, Wang, Yang, Wang, Liu, Jiang, and Zhong]{hunyuanvideo}
Weijie Kong, Qi Tian, Zijian Zhang, Rox Min, Zuozhuo Dai, Jin Zhou, Jiangfeng Xiong, Xin Li, Bo Wu, Jianwei Zhang, Kathrina Wu, Qin Lin, Junkun Yuan, Yanxin Long, Aladdin Wang, Andong Wang, Changlin Li, Duojun Huang, Fang Yang, Hao Tan, Hongmei Wang, Jacob Song, Jiawang Bai, Jianbing Wu, Jinbao Xue, Joey Wang, Kai Wang, Mengyang Liu, Pengyu Li, Shuai Li, Weiyan Wang, Wenqing Yu, Xinchi Deng, Yang Li, Yi Chen, Yutao Cui, Yuanbo Peng, Zhentao Yu, Zhiyu He, Zhiyong Xu, Zixiang Zhou, Zunnan Xu, Yangyu Tao, Qinglin Lu, Songtao Liu, Dax Zhou, Hongfa Wang, Yong Yang, Di Wang, Yuhong Liu, Jie Jiang, and Caesar Zhong.
\newblock Hunyuanvideo: A systematic framework for large video generative models, 2025.

\bibitem[Ku et~al.(2024)Ku, Wei, Ren, Yang, and Chen]{anyv2v}
Max Ku, Cong Wei, Weiming Ren, Huan Yang, and Wenhu Chen.
\newblock Anyv2v: A plug-and-play framework for any videoto-video editing tasks.
\newblock \emph{arXiv preprint arXiv:2403.14468}, 2\penalty0 (3):\penalty0 5, 2024.

\bibitem[Liu et~al.(2024)Liu, Li, Zhang, Lan, and Liu]{stablev2v}
Chang Liu, Rui Li, Kaidong Zhang, Yunwei Lan, and Dong Liu.
\newblock Stablev2v: Stablizing shape consistency in video-to-video editing.
\newblock \emph{arXiv preprint arXiv:2411.11045}, 2024.

\bibitem[Liu et~al.(2025)Liu, Wang, Wang, Liu, Zhang, Lee, Li, Yu, Lin, Kim, et~al.]{genprop}
Shaoteng Liu, Tianyu Wang, Jui-Hsien Wang, Qing Liu, Zhifei Zhang, Joon-Young Lee, Yijun Li, Bei Yu, Zhe Lin, Soo~Ye Kim, et~al.
\newblock Generative video propagation.
\newblock In \emph{Proceedings of the Computer Vision and Pattern Recognition Conference}, pages 17712--17722, 2025.

\bibitem[Loshchilov and Hutter(2017)]{loshchilov2017adamw}
Ilya Loshchilov and Frank Hutter.
\newblock Decoupled weight decay regularization.
\newblock \emph{arXiv preprint arXiv:1711.05101}, 2017.

\bibitem[Ma et~al.(2025)Ma, Liu, Zhu, Yang, Feng, Zhang, Li, Han, Qi, and Chen]{ma2025follow}
Yue Ma, Yulong Liu, Qiyuan Zhu, Ayden Yang, Kunyu Feng, Xinhua Zhang, Zhifeng Li, Sirui Han, Chenyang Qi, and Qifeng Chen.
\newblock Follow-your-motion: Video motion transfer via efficient spatial-temporal decoupled finetuning.
\newblock \emph{arXiv preprint arXiv:2506.05207}, 2025.

\bibitem[Ouyang et~al.(2024{\natexlab{a}})Ouyang, Dong, Yang, Si, and Pan]{i2vedit}
Wenqi Ouyang, Yi Dong, Lei Yang, Jianlou Si, and Xingang Pan.
\newblock I2vedit: First-frame-guided video editing via image-to-video diffusion models.
\newblock In \emph{SIGGRAPH Asia 2024 Conference Papers}, pages 1--11, 2024{\natexlab{a}}.

\bibitem[Ouyang et~al.(2024{\natexlab{b}})Ouyang, Dong, Yang, Si, and Pan]{ouyang2024i2vedit}
Wenqi Ouyang, Yi Dong, Lei Yang, Jianlou Si, and Xingang Pan.
\newblock I2vedit: First-frame-guided video editing via image-to-video diffusion models.
\newblock In \emph{SIGGRAPH Asia 2024 Conference Papers}, pages 1--11, 2024{\natexlab{b}}.

\bibitem[Peebles and Xie(2023)]{dit}
William Peebles and Saining Xie.
\newblock Scalable diffusion models with transformers.
\newblock In \emph{Proceedings of the IEEE/CVF international conference on computer vision}, pages 4195--4205, 2023.

\bibitem[Qu et~al.(2025)Qu, Zhang, Liu, Wang, Jiang, Gao, Ye, Du, Yuan, and Wu]{qu2025tokenflow}
Liao Qu, Huichao Zhang, Yiheng Liu, Xu Wang, Yi Jiang, Yiming Gao, Hu Ye, Daniel~K Du, Zehuan Yuan, and Xinglong Wu.
\newblock Tokenflow: Unified image tokenizer for multimodal understanding and generation.
\newblock In \emph{Proceedings of the Computer Vision and Pattern Recognition Conference}, pages 2545--2555, 2025.

\bibitem[Ren et~al.(2024)Ren, Liu, Zeng, Lin, Li, Cao, Chen, Huang, Chen, Yan, et~al.]{groundsam}
Tianhe Ren, Shilong Liu, Ailing Zeng, Jing Lin, Kunchang Li, He Cao, Jiayu Chen, Xinyu Huang, Yukang Chen, Feng Yan, et~al.
\newblock Grounded sam: Assembling open-world models for diverse visual tasks.
\newblock \emph{arXiv preprint arXiv:2401.14159}, 2024.

\bibitem[Su et~al.(2024)Su, Ahmed, Lu, Pan, Bo, and Liu]{su2024rope}
Jianlin Su, Murtadha Ahmed, Yu Lu, Shengfeng Pan, Wen Bo, and Yunfeng Liu.
\newblock Roformer: Enhanced transformer with rotary position embedding.
\newblock \emph{Neurocomputing}, 568:\penalty0 127063, 2024.

\bibitem[Team(2024)]{LucyEdit2024}
Decart~AI Team.
\newblock Lucy-edit: Open-weight text-guided video editing.
\newblock Technical report, Lucy-Edit Team, 2024.
\newblock Accessed: 2025-10-28.

\bibitem[Team(2025)]{wan2.1}
Wan Team.
\newblock Wan: Open and advanced large-scale video generative models.
\newblock 2025.

\bibitem[Wang et~al.(2024)Wang, Bai, Tan, Wang, Fan, Bai, Chen, Liu, Wang, Ge, et~al.]{qwen2}
Peng Wang, Shuai Bai, Sinan Tan, Shijie Wang, Zhihao Fan, Jinze Bai, Keqin Chen, Xuejing Liu, Jialin Wang, Wenbin Ge, et~al.
\newblock Qwen2-vl: Enhancing vision-language model's perception of the world at any resolution.
\newblock \emph{arXiv preprint arXiv:2409.12191}, 2024.

\bibitem[Wang et~al.(2025{\natexlab{a}})Wang, Shi, Ou, Chen, Lin, Wang, Jiang, Yang, Zheng, Tao, et~al.]{koala}
Qiuheng Wang, Yukai Shi, Jiarong Ou, Rui Chen, Ke Lin, Jiahao Wang, Boyuan Jiang, Haotian Yang, Mingwu Zheng, Xin Tao, et~al.
\newblock Koala-36m: A large-scale video dataset improving consistency between fine-grained conditions and video content.
\newblock In \emph{Proceedings of the Computer Vision and Pattern Recognition Conference}, pages 8428--8437, 2025{\natexlab{a}}.

\bibitem[Wang et~al.(2025{\natexlab{b}})Wang, Liu, Lin, Liu, Yi, Wang, and Ma]{omnistyle}
Ye Wang, Ruiqi Liu, Jiang Lin, Fei Liu, Zili Yi, Yilin Wang, and Rui Ma.
\newblock Omnistyle: Filtering high quality style transfer data at scale.
\newblock In \emph{Proceedings of the Computer Vision and Pattern Recognition Conference}, pages 7847--7856, 2025{\natexlab{b}}.

\bibitem[Wang et~al.(2025{\natexlab{c}})Wang, Wang, Ma, Hu, Xu, and Guo]{wang2025videodirector}
Yukun Wang, Longguang Wang, Zhiyuan Ma, Qibin Hu, Kai Xu, and Yulan Guo.
\newblock Videodirector: Precise video editing via text-to-video models.
\newblock In \emph{Proceedings of the Computer Vision and Pattern Recognition Conference}, pages 2589--2598, 2025{\natexlab{c}}.

\bibitem[Wu et~al.(2025{\natexlab{a}})Wu, Li, Zhou, Lin, Gao, Yan, ming Yin, Bai, Xu, Chen, Chen, Tang, Zhang, Wang, Yang, Yu, Cheng, Liu, Li, Zhang, Meng, Wei, Ni, Chen, Cao, Peng, Qu, Wu, Wang, Yu, Wen, Feng, Xu, Wang, Zhang, Zhu, Wu, Cai, and Liu]{qwenimage}
Chenfei Wu, Jiahao Li, Jingren Zhou, Junyang Lin, Kaiyuan Gao, Kun Yan, Sheng ming Yin, Shuai Bai, Xiao Xu, Yilei Chen, Yuxiang Chen, Zecheng Tang, Zekai Zhang, Zhengyi Wang, An Yang, Bowen Yu, Chen Cheng, Dayiheng Liu, Deqing Li, Hang Zhang, Hao Meng, Hu Wei, Jingyuan Ni, Kai Chen, Kuan Cao, Liang Peng, Lin Qu, Minggang Wu, Peng Wang, Shuting Yu, Tingkun Wen, Wensen Feng, Xiaoxiao Xu, Yi Wang, Yichang Zhang, Yongqiang Zhu, Yujia Wu, Yuxuan Cai, and Zenan Liu.
\newblock Qwen-image technical report, 2025{\natexlab{a}}.

\bibitem[Wu et~al.(2025{\natexlab{b}})Wu, Chen, Li, Wang, Xie, and Zhang]{insvie}
Yuhui Wu, Liyi Chen, Ruibin Li, Shihao Wang, Chenxi Xie, and Lei Zhang.
\newblock Insvie-1m: Effective instruction-based video editing with elaborate dataset construction.
\newblock In \emph{Proceedings of the IEEE/CVF International Conference on Computer Vision}, pages 16692--16701, 2025{\natexlab{b}}.

\bibitem[Xi et~al.(2025)Xi, Yang, Zhao, Xu, Li, Li, Lin, Cai, Zhang, Li, et~al.]{xi2025sparse}
Haocheng Xi, Shuo Yang, Yilong Zhao, Chenfeng Xu, Muyang Li, Xiuyu Li, Yujun Lin, Han Cai, Jintao Zhang, Dacheng Li, et~al.
\newblock Sparse videogen: Accelerating video diffusion transformers with spatial-temporal sparsity.
\newblock \emph{arXiv preprint arXiv:2502.01776}, 2025.

\bibitem[Yang et~al.(2024)Yang, Kang, Huang, Zhao, Xu, Feng, and Zhao]{depthanything}
Lihe Yang, Bingyi Kang, Zilong Huang, Zhen Zhao, Xiaogang Xu, Jiashi Feng, and Hengshuang Zhao.
\newblock Depth anything v2.
\newblock \emph{Advances in Neural Information Processing Systems}, 37:\penalty0 21875--21911, 2024.

\bibitem[Yang et~al.(2025)Yang, Teng, Zheng, Ding, Huang, Xu, Yang, Hong, Zhang, Feng, Yin, Yuxuan.Zhang, Wang, Cheng, Xu, Gu, Dong, and Tang]{cogvideox}
Zhuoyi Yang, Jiayan Teng, Wendi Zheng, Ming Ding, Shiyu Huang, Jiazheng Xu, Yuanming Yang, Wenyi Hong, Xiaohan Zhang, Guanyu Feng, Da Yin, Yuxuan.Zhang, Weihan Wang, Yean Cheng, Bin Xu, Xiaotao Gu, Yuxiao Dong, and Jie Tang.
\newblock Cogvideox: Text-to-video diffusion models with an expert transformer.
\newblock In \emph{The Thirteenth International Conference on Learning Representations}, 2025.

\bibitem[Yatim et~al.(2024)Yatim, Fridman, Bar-Tal, Kasten, and Dekel]{yatim2024stdf}
Danah Yatim, Rafail Fridman, Omer Bar-Tal, Yoni Kasten, and Tali Dekel.
\newblock Space-time diffusion features for zero-shot text-driven motion transfer.
\newblock In \emph{Proceedings of the IEEE/CVF Conference on Computer Vision and Pattern Recognition}, pages 8466--8476, 2024.

\bibitem[Ye et~al.(2025)Ye, He, Liu, Wang, Wang, Wan, Zhang, Gai, Chen, and Luo]{unic}
Zixuan Ye, Xuanhua He, Quande Liu, Qiulin Wang, Xintao Wang, Pengfei Wan, Di Zhang, Kun Gai, Qifeng Chen, and Wenhan Luo.
\newblock Unic: Unified in-context video editing, 2025.

\bibitem[Yoon et~al.(2025)Yoon, Yu, and Bansal]{raccoon}
Jaehong Yoon, Shoubin Yu, and Mohit Bansal.
\newblock Raccoon: Versatile instructional video editing with auto-generated narratives.
\newblock In \emph{Proceedings of the 2025 Conference on Empirical Methods in Natural Language Processing}, pages 27960--27996, 2025.

\bibitem[Zhang et~al.(2025)Zhang, Liao, Zhang, Meng, Wan, Yan, and Cheng]{zhang2025videorepa}
Xiangdong Zhang, Jiaqi Liao, Shaofeng Zhang, Fanqing Meng, Xiangpeng Wan, Junchi Yan, and Yu Cheng.
\newblock Videorepa: Learning physics for video generation through relational alignment with foundation models.
\newblock \emph{arXiv preprint arXiv:2505.23656}, 2025.

\bibitem[Zhang et~al.(2024)Zhang, Dai, Qin, and Wang]{effived}
Zhenghao Zhang, Zuozhuo Dai, Long Qin, and Weizhi Wang.
\newblock Effived:efficient video editing via text-instruction diffusion models, 2024.

\bibitem[Zi et~al.(2025)Zi, Ruan, Chen, Qi, Hao, Zhao, Huang, Liang, Xiao, and Wong]{senorita}
Bojia Zi, Penghui Ruan, Marco Chen, Xianbiao Qi, Shaozhe Hao, Shihao Zhao, Youze Huang, Bin Liang, Rong Xiao, and Kam-Fai Wong.
\newblock Se\~norita-2m: A high-quality instruction-based dataset for general video editing by video specialists, 2025.

\end{thebibliography}
}

\newpage

\section{Additional Information about FFP-300K}
The video frames shown in our figures have been packaged and uploaded. Please refer to the accompanying zip file for details.
\subsection{Dataset Construction}
\paragraph{Prompts used.} Our data construction pipeline follows a two-track modular pipeline and we use Qwen2.5-VL-72B-Instruct~\cite{bai2025qwen2.5vl} to produce the prompts for both local editing and global stylization.

\subsubsection{Local Editing}
To identify the primary editable objects in each video, we use a prompt that analyzes the first frame.
\begin{tcolorbox}[myprompt]
\textbf{Object Identification.}  \\
You are given a single video frame. Identify the main editable object in this frame.\\
\noindent Rules:
\begin{itemize}[leftmargin=1.5em]
    \item Output THREE lowercase category word only (e.g., person, car, dog, ball, cup, bottle, phone, bag, plant, flower, sign, tableware).
    \item Do not describe background or actions.
    \item If several candidates exist, choose the smallest salient object that humans often edit/remove (e.g., ball before person in sports; cup before hands on a table).
\end{itemize}
Output strictly in JSON: \{``Object'': ``Category''\}
\end{tcolorbox}

\noindent The original caption is constructed to preserve the scene context and serve as reference when replacing the masked object for swap tasks.
\begin{tcolorbox}[myprompt]
\textbf{Original Caption.} \\
You are given a short video.
Write ONE concise caption in present tense (18–30 words) describing only the stable scene elements (location, background, persistent subject categories).
\noindent Rules:
\begin{itemize}[leftmargin=1.5em]
    \item Describe what remains visually consistent across the clip.
    \item Use generic categories for moving actors (e.g., ``a person'', ``three red cups on a white table'', ``a yellow taxi by a street'').
    \item Avoid counts unless they are constant; avoid names, brands, emotions, camera terms.
    \item No negations (e.g., ``no/without'').
    \item Keep it objective, concrete, and free of text/letters.
\end{itemize}
Output strictly in JSON: {``Caption'': ``The video shows ...'' }
\end{tcolorbox}

\noindent The removal caption describes the scene without the target object identified earlier.
\begin{tcolorbox}[myprompt]
\textbf{Removal Caption.} \\
You are given an original video caption and a target object to remove. Rewrite the caption so it naturally describes the scene as if that object never existed.
\noindent Rules:
\begin{itemize}[leftmargin=1.5em]
    \item Output ONE fluent sentence in present tense, 35–60 words.
    \item Start with: `` The video shows ... ''
    \item Do NOT mention the removed object, any pronouns referring to it, or actions tied to it (e.g., holding, touching, pouring).
    \item  Do NOT use negations like ``no/without''.
    \item Do NOT invent new objects or text/letters that the original background did not imply.
    \item Keep only concrete, persistent background elements (location, surfaces, vehicles, trees, buildings, sky, lighting, colors, furniture).
\end{itemize}
Input:  {``Original\_Caption'': ``Original\_Caption'', ``Removal'':``Object''} \\
Output strictly in JSON: \{``Remove\_Caption'': ``The video shows ... ''\}
\end{tcolorbox}

\noindent For swap tasks, this prompt determines whether the edited video should be classified as a swap or a modification.
\begin{tcolorbox}[myprompt]
\textbf{Task Discrimination. } \\
You are given two short videos:
\begin{itemize}
    \item Video A: Source Video
    \item Video B: Generated after editing
\end{itemize}
\noindent Task: Compare A and B by examining the first, middle, and last frames.
\begin{itemize}
    \item Decide:
    \begin{itemize}
        \item swap: an object in A is replaced by a different object in B. The region still contains an object, but its identity changes.
        \item modification: the same object remains in B, but its attributes (shape, color, texture, style, size, letters, patterns, or fine details) are edited without replacing it with a different object.
    \end{itemize}
\end{itemize}
If the object’s identity clearly changes, classify as ``swap.
If only attributes or features change while the object stays the same, classify as ``modification.

Output strictly in JSON: \{``Task'': ``Swap''\} or \{``Task'': ``Modification''\}
\end{tcolorbox}

\subsubsection{Global Stylization}
For global stylization, a cinematic caption is first constructed to summarize the scene and atmosphere of the input artistic image for source video generation.
\begin{tcolorbox}[myprompt]
\textbf{Image-to-Video Caption.} You are an image-to-video prompt generator.   Analyze the input image \{image\} and output only one cinematic video prompt.  
\noindent Rules:
\begin{itemize}[leftmargin=1.5em]
    \item Provide a concise scene description (environment, atmosphere, subjects). 
    \item Do not add new motions to the subjects; keep them static as in the image.  
    \item Focus on cinematic camera work: wide shots, dolly-in, pans, or close-ups.  
    \item You may suggest smooth transitions or scene framing, but no new actions.  
    \item Limit the output to 3–4 sentences. 
\end{itemize}
Output: only the final video prompt.
\end{tcolorbox}

\noindent For stylized video generation, the following prompt produces a detailed style description based on both the reference style image and the source video.
\begin{tcolorbox}[myprompt]
\textbf{Style Caption.} Apply style transfer using the reference image, but keep the output cinematic and natural.
\noindent Rules:
\begin{itemize}[leftmargin=1.5em]
    \item Style Control: Use soft, balanced colors with reduced saturation, no overexposure.
    \item Subject Preservation: Preserve the subject’s natural tones and details (do not oversaturate). 
    \item Lighting \& Texture: Maintain subtle textures, soft lighting, and a film-like atmosphere. 
    \item Constraints: Avoid harsh highlights, neon effects, or unnatural color shifts. 
\end{itemize}
\end{tcolorbox}

\subsection{Dataset Analysis}

\begin{figure}
    \centering
    \includegraphics[width=\linewidth]{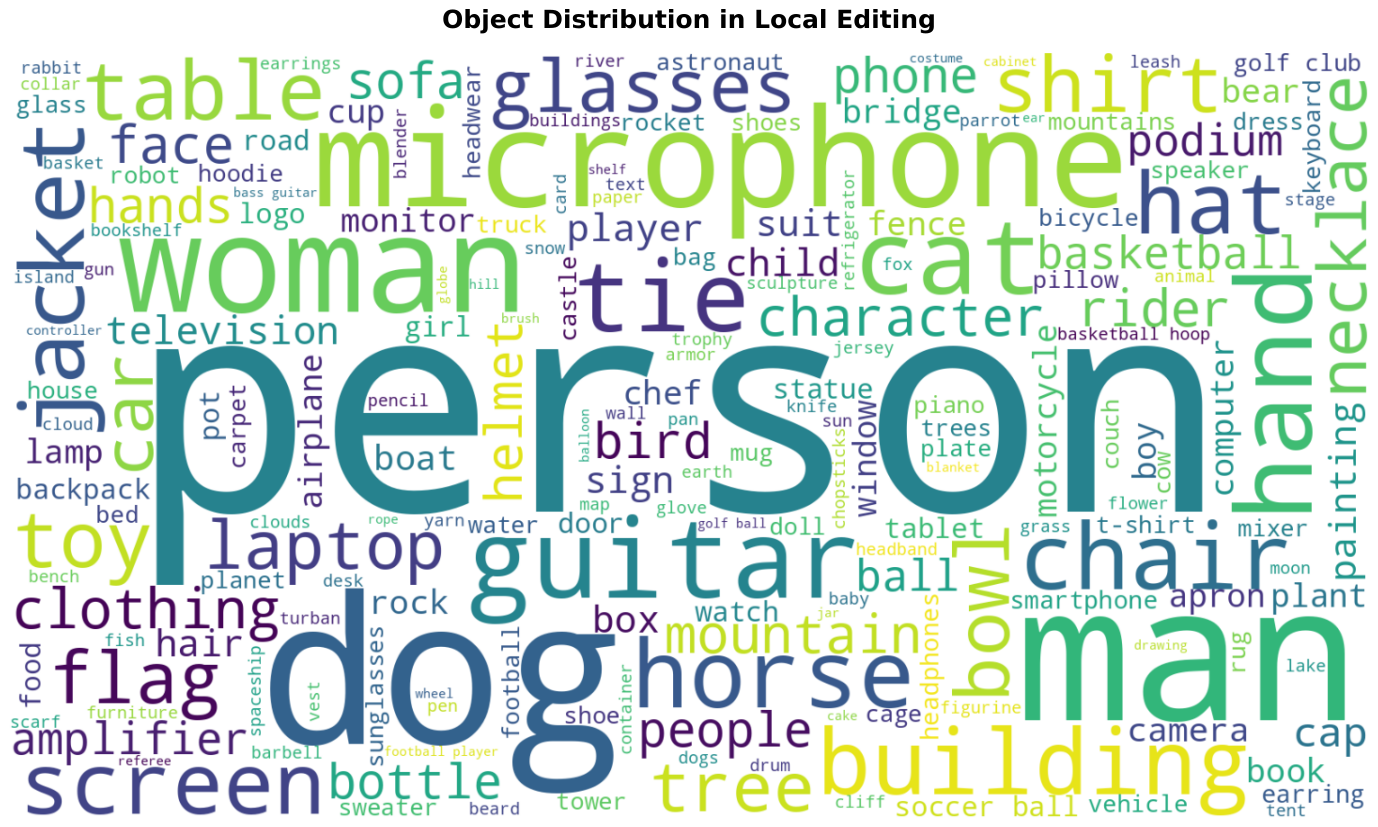}
    \caption{Word cloud of edited objects of the local editing subst of FFP-300K.}
    \label{fig:wordcloud}
\end{figure}

\paragraph{Distribution of edited objects.} We visualize the objects selected for local editing in FFP-300K in Fig.~\ref{fig:wordcloud}. The word cloud highlights substantial diversity in the edited-object space of the local-editing subset of FFP-300K: while people and hand-held items (e.g., person, microphone, guitar, phone) are prominent, there is a wide spread of categories spanning furniture and electronics (table, chair, laptop), animals and nature (horse, tree, bird), vehicles and buildings, and many everyday objects. This long-tailed, semantically rich distribution indicates the dataset supports a broad range of local-editing scenarios, from fine-grained human-centric manipulations to structurally complex scene elements. Consequently, models trained on FFP-300K are exposed to varied object types and contexts, which helps foster robustness and generalization across diverse editing tasks. 

\begin{figure}
    \centering
    \includegraphics[width=\linewidth]{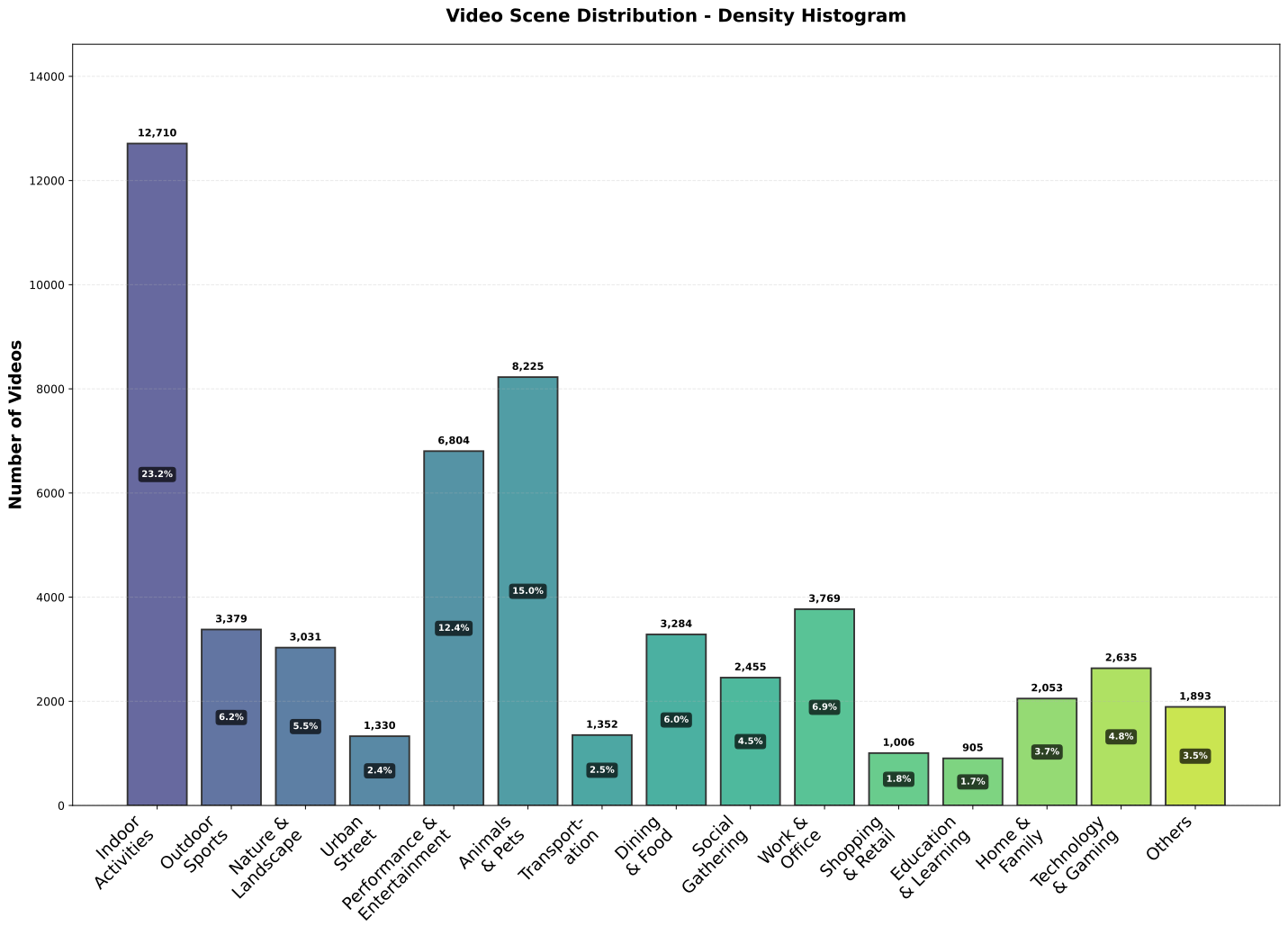}
    \caption{Scene distribution of the local editing subset of FFP-300K}
    \label{fig:scene}
\end{figure}

\paragraph{Distribution of video content.} To demonstrate the content diversity of FFP-300K, for each source video adopted, we extract 5 frames and ask Qwen2.5-VL to classify them into 15 predefined scenes, of which the distribution is shown in Fig.~\ref{fig:scene}. The scene distribution of the local-editing subset is strongly skewed toward a few dominant contexts—Indoor Activities (12,710 videos, ~29.2\%), Performance \& Entertainment (8,225, ~19.0\%) and Urban Scenes (6,904, ~15.9\%)—while the remaining categories (e.g., Outdoor Activities, Sports, Nature, Animals, Transport, Technology, Medical, etc.) form a long tail with individual shares typically below 8\%. This composition provides dense coverage of common indoor and urban editing scenarios that are crucial for real-world applications, while still retaining broad scene diversity for generalization.

\subsection{Visualization of FFP-300K}
To illustrate the visual results of FFP-300K, we provide representative examples from the two tracks of our data construction pipeline. Both tracks maintain spatial coherence and temporal consistency across all frames, enabling the model to learn strong motion priors through the first-frame propagation paradigm and supporting reliable video editing.

\begin{figure*}[t]
  \centering
  \includegraphics[width=0.95\textwidth]{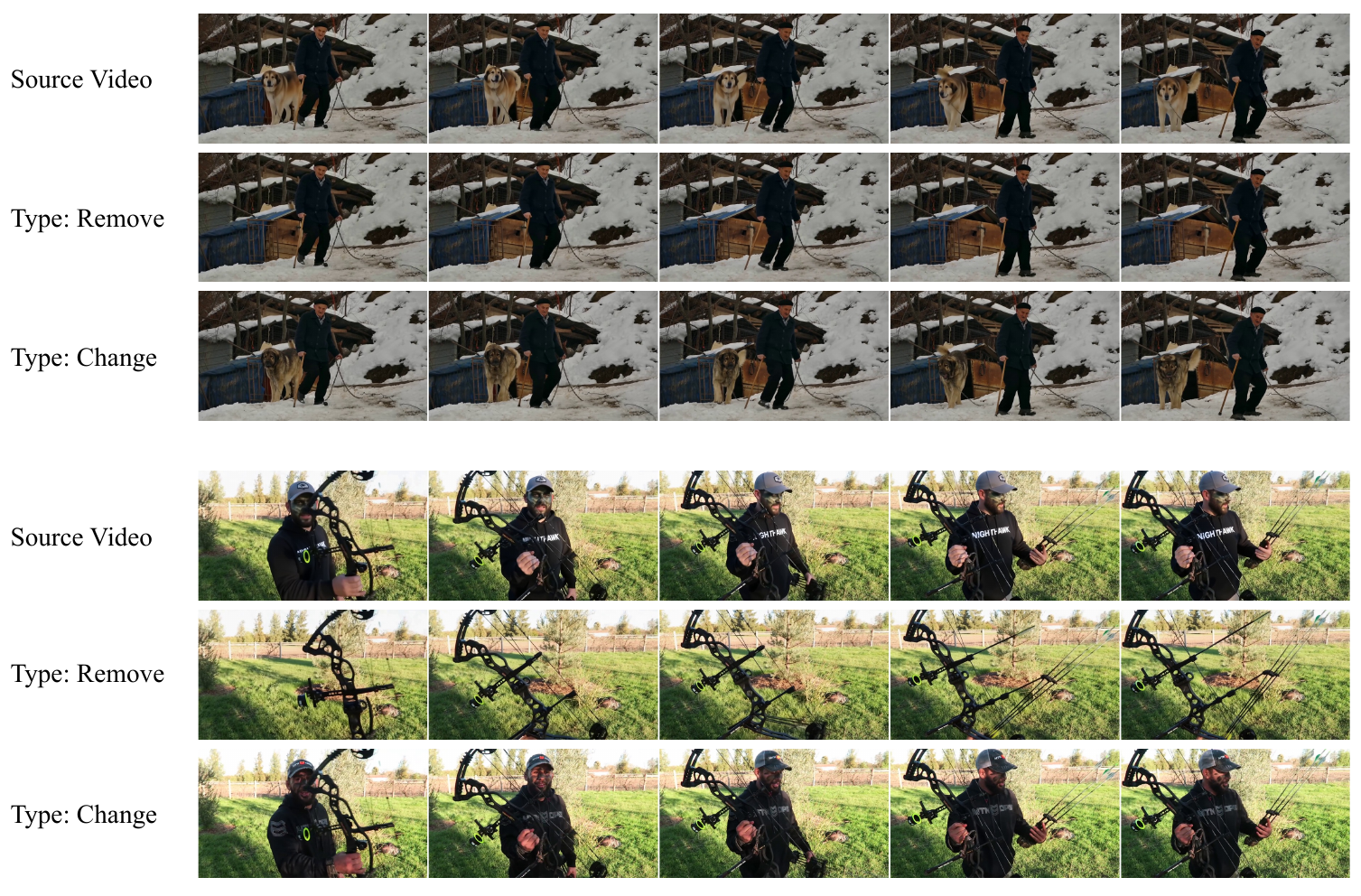}
  \vspace{-0.2in}
  \caption{ The visualization of local editing track in FFP-300K.
    }
  \label{fig:sup_swap&remove_demo}
\end{figure*}

\paragraph{Local Editing.} The local editing track constructs object-level samples using remove and swap manipulations. These samples are generated by editing specific target objects in the source video while keeping the surrounding scene unchanged, forming paired sequences that cover diverse object categories and scene contexts. As shown in Fig.~\ref{fig:sup_swap&remove_demo}, these examples reflect the broad coverage of fine-grained object manipulations and varied local-editing scenarios present in FFP-300K.

\begin{figure*}[t]
  \centering
  \includegraphics[width=0.95\textwidth]{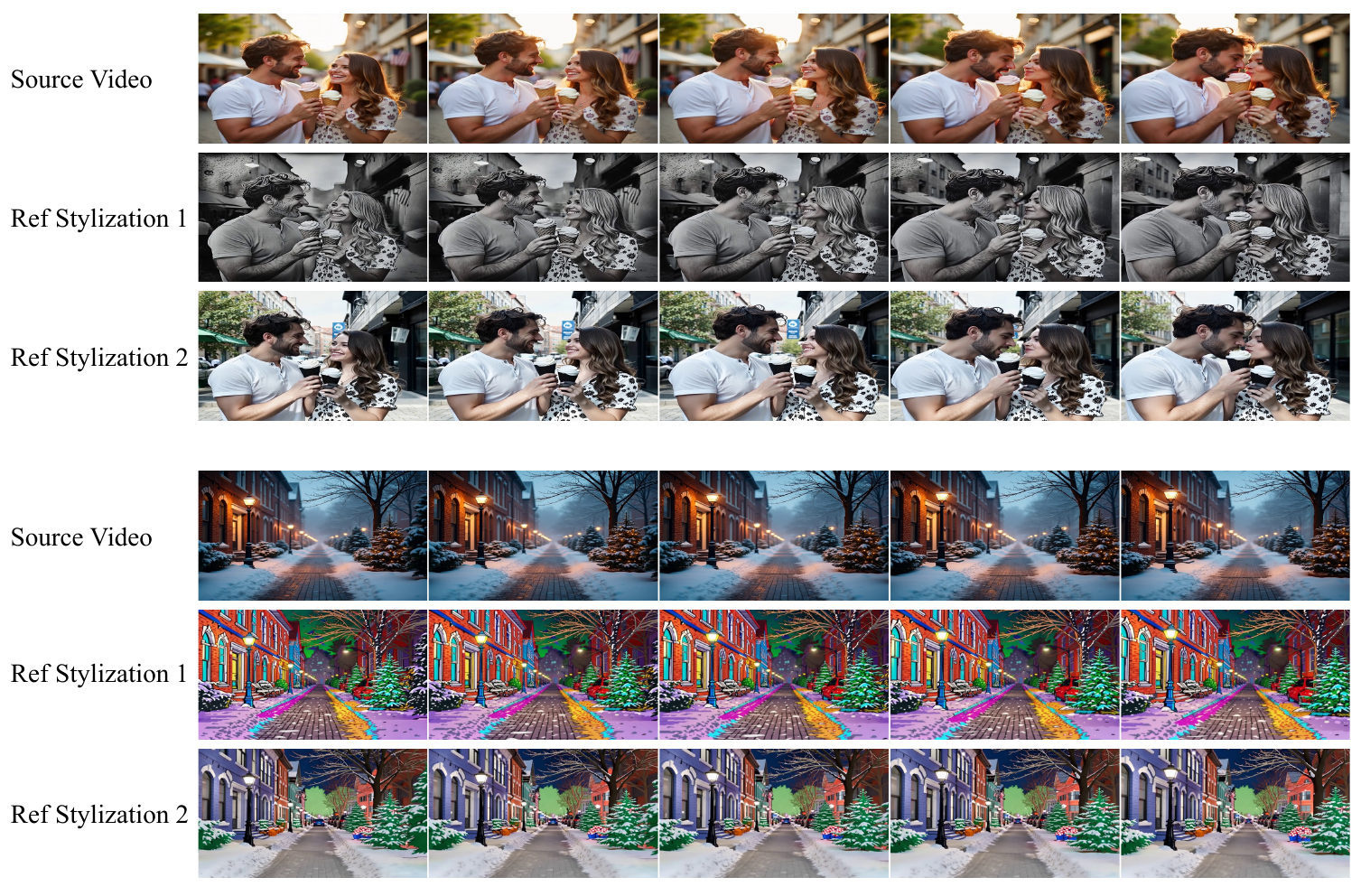}
  \vspace{-0.2in}
  \caption{ The visualization of global stylization track in FFP-300K.}
  \vspace{-0.2in}
  \label{fig:sup_stylization_demo}
\end{figure*}

\paragraph{Global Stylization.} The global stylization track generates full-scene style-transfer samples by applying the appearance of a reference image to the entire source video. Each source video is paired with multiple reference images, producing multiple stylized sequences that span a wide range of aesthetic styles. As illustrated in Fig.~\ref{fig:sup_stylization_demo}, these samples expand the appearance diversity of the dataset and represent the full-scene stylization capabilities captured in FFP-300K.

\section{Additional Method Details}
\label{sec:supp_attention_classification}

\subsection{Attention Head Classification Heuristic}
Our proposed AST-RoPE requires pre-classification for each self-attention head. While previous methods such as SparseVidGen and Follow-your-motion utilize sample-specific classification, we find that the category of each attention head is generally sample-agnostic, which is intuitively reasonable that each head learns fixed prior knowledge. Therefore we design a simple classification strategy as follows.

\paragraph{Grid-based Partitioning of the Attention Map.}
For a given self-attention head and an input video with $F$ frames, each of resolution $H \times W$, the total number of tokens is $N = F' \times H' \times W'$. The attention map is a matrix $\mathbf{A} \in \mathbb{R}^{N \times N}$. We conceptually partition this large matrix into a $F' \times F'$ grid of smaller sub-matrices. Each sub-matrix $\mathbf{A}_{ij}$ represents the attention from all tokens in the source latent frame $i$ to all tokens in the target latent frame $j$.

\paragraph{Quantifying Attention Density.}
We measure the ``activity'' within each grid by calculating its attention density. The attention density $\rho_{ij}$ for a grid $\mathbf{A}_{ij}$ is defined as the proportion of its elements that are non-zero. In practice, due to the softmax function, all attention scores are positive. We therefore define density as the proportion of attention scores exceeding a small threshold $\epsilon$ (e.g., $\epsilon = 10^{-6}$) to filter out negligible floating-point values.
\begin{equation}
    \rho_{ij} = \frac{1}{H \times W \times H \times W} \sum_{u=1}^{HW} \sum_{v=1}^{HW} \mathbb{I}(\mathbf{A}_{ij}[u, v] > \epsilon)
\end{equation}
where $\mathbb{I}(\cdot)$ is the indicator function.

\paragraph{The Classification Rule.}
Our heuristic compares the strongest temporal signal against the weakest spatial signal. Let $\mathcal{D}_{\text{diag}} = \{ \rho_{ii} \mid i \in [1, F'] \}$ be the set of densities for all diagonal (spatial) grids, and $\mathcal{D}_{\text{non-diag}} = \{ \rho_{ij} \mid i, j \in [1, F'], i \neq j \}$ be the set for all non-diagonal (temporal) grids.

An attention head is classified as \textbf{temporal} if its maximum non-diagonal attention density is greater than its minimum diagonal attention density. Otherwise, it is classified as \textbf{spatial}.
\begin{equation}
\text{Head Type} = 
\begin{cases} 
\text{Temporal} & \text{if } \max(\mathcal{D}_{\text{non-diag}}) > \min(\mathcal{D}_{\text{diag}}) \\
\text{Spatial} & \text{otherwise}
\end{cases}
\end{equation}
The intuition is that for a head to be genuinely temporal, its cross-frame attention must be meaningful and stronger than its most diffuse, weakest intra-frame attention. A head that only pays weak, noisy attention across frames but strong attention within frames will be correctly classified as spatial.

\paragraph{Final Classification via Majority Voting.}
The behavior of an attention head can be content-dependent. To obtain a stable and generalizable classification, we do not rely on a single video sample. Instead, we apply the classification process described above to a set of \textbf{10 diverse video samples} randomly drawn from our validation set. The final, definitive classification for each attention head is determined by a \textbf{majority vote} on the outcomes from these 10 samples. This aggregation ensures that the assigned role reflects the head's typical behavior rather than an artifact of a specific input.

\section{Additional Experiment Results}
\subsection{Experiments on UNICBench}

\begin{figure*}[t]
  \centering
  \includegraphics[width=1\textwidth]{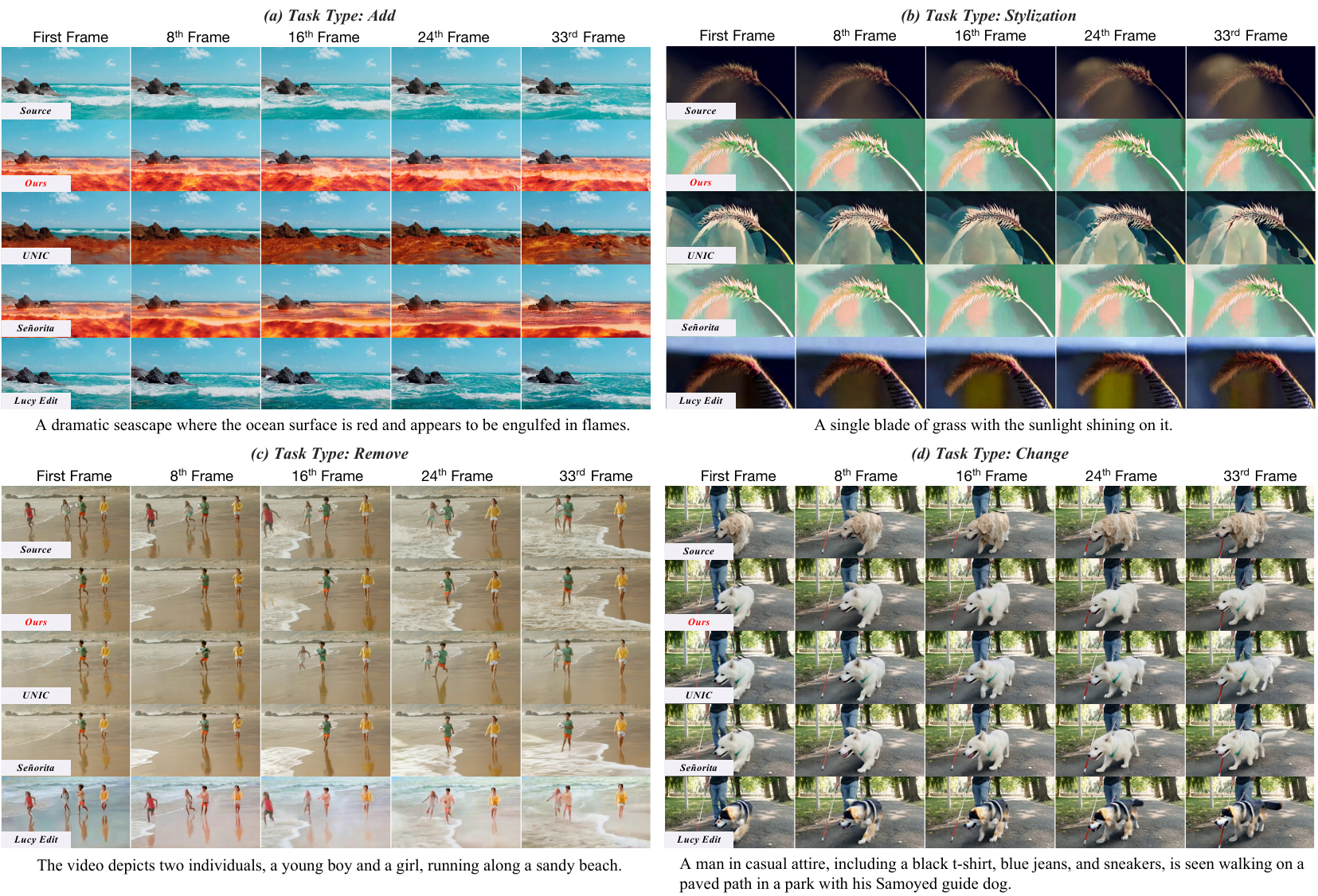}
  \vspace{-0.2in}
  \caption{\textbf{Qualitative Comparison.} We choose top three methods in quantitative comparison to compare with our-33f visual results across local editing and global stylization tasks.}
  \vspace{-0.2in}
  \label{fig:qualitative_unic}
\end{figure*}

As a supplement to the experiment in the main paper, we further conduct experiments on UNICBench~\cite{unic}, which is filtered by us with the same principle as for EditVerseBench to delete samples that are not suitable for FFP. The whole test set contains 128 videos, covering tasks of add, delete, change and stylization. We adopt UNIC~\cite{unic}, AnyV2V~\cite{anyv2v}, LucyEdit~\cite{LucyEdit2024} and Senorita~\cite{senorita} as baseline methods, among which the results of UNIC and AnyV2V are provided by UNIC, and results of the other two methods are produced by us. We adopt the same metrics as EditVerseBench, which are presented in Tab.~\ref{tab:quantitative_unic}. Our method receives the best performance in terms of all metrics. The qualitative comparison is shown in Fig.~\ref{fig:qualitative_unic}, which further demonstrates that our method is not only more accurate for editing but also visually better.

\begin{table}[t]
\small
\centering
\resizebox{1\linewidth}{!}{
\begin{tabular}{c c c c c c c}
\toprule
 & \multicolumn{2}{c}{\textbf{Temporal Consistency}} & \multicolumn{2}{c}{\textbf{Text Alignment}} & \textbf{Video Quality} & \textbf{VLM Evaluation} \\
Type & CLIP $\uparrow$ & DINO $\uparrow$ & Frame $\uparrow$ & Video $\uparrow$ & Pick Score $\uparrow$ & VLM Score $\uparrow$ \\
\midrule
AnyV2V &  0.941 & 	0.92 & 23.597 & 20.138 & 19.864 & 4.132 \\
LucyEdit & 0.978 & 0.977 & 22.171 & 18.036 & 19.612 & 5.065 \\
Senorita & 0.985 & 0.981 & 24.197 & 20.273 & 19.950 & 6.648 \\
UNIC & 0.980 & 0.973 & 24.267 & 20.116 & 19.182 & 5.203 \\
\rowcolor{gray!20} \cellcolor{white} \textbf{Ours} & \textbf{0.986} & \textbf{0.982} & \textbf{24.879} & \textbf{20.733} & \textbf{19.951} & \textbf{6.672} \\
\bottomrule
\end{tabular}
}
\vspace{-0.1in}
\caption{\textbf{Quantitative comparison.} We compared three types of video editing methods on UNICBench. The best results are highlighted in \textbf{bold}.  
}
\label{tab:quantitative_unic}
\vspace{-0.2in}
\end{table}


\subsection{More Results on EditVerseBench}
\begin{figure*}[p]
  \centering
  \includegraphics[height=1.4\textwidth]{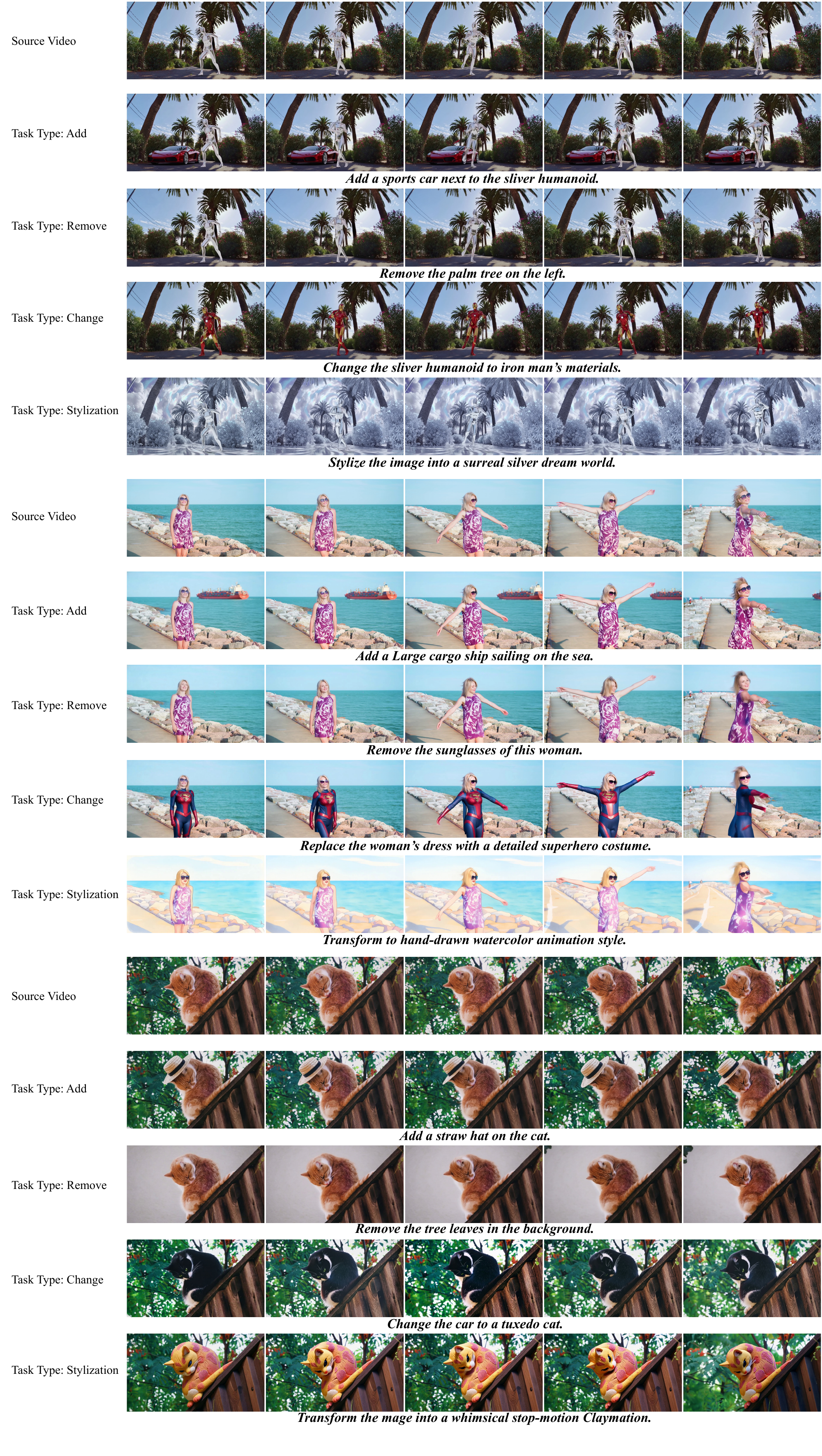}
  \vspace{-0.2in}
  \caption{More results of local editing and global stylization tasks on EditVerseBench.}
  \label{fig:sup_fulltask_editversebench}
\end{figure*}

\begin{figure*}[t]
  \centering
  \includegraphics[width=0.95\textwidth]{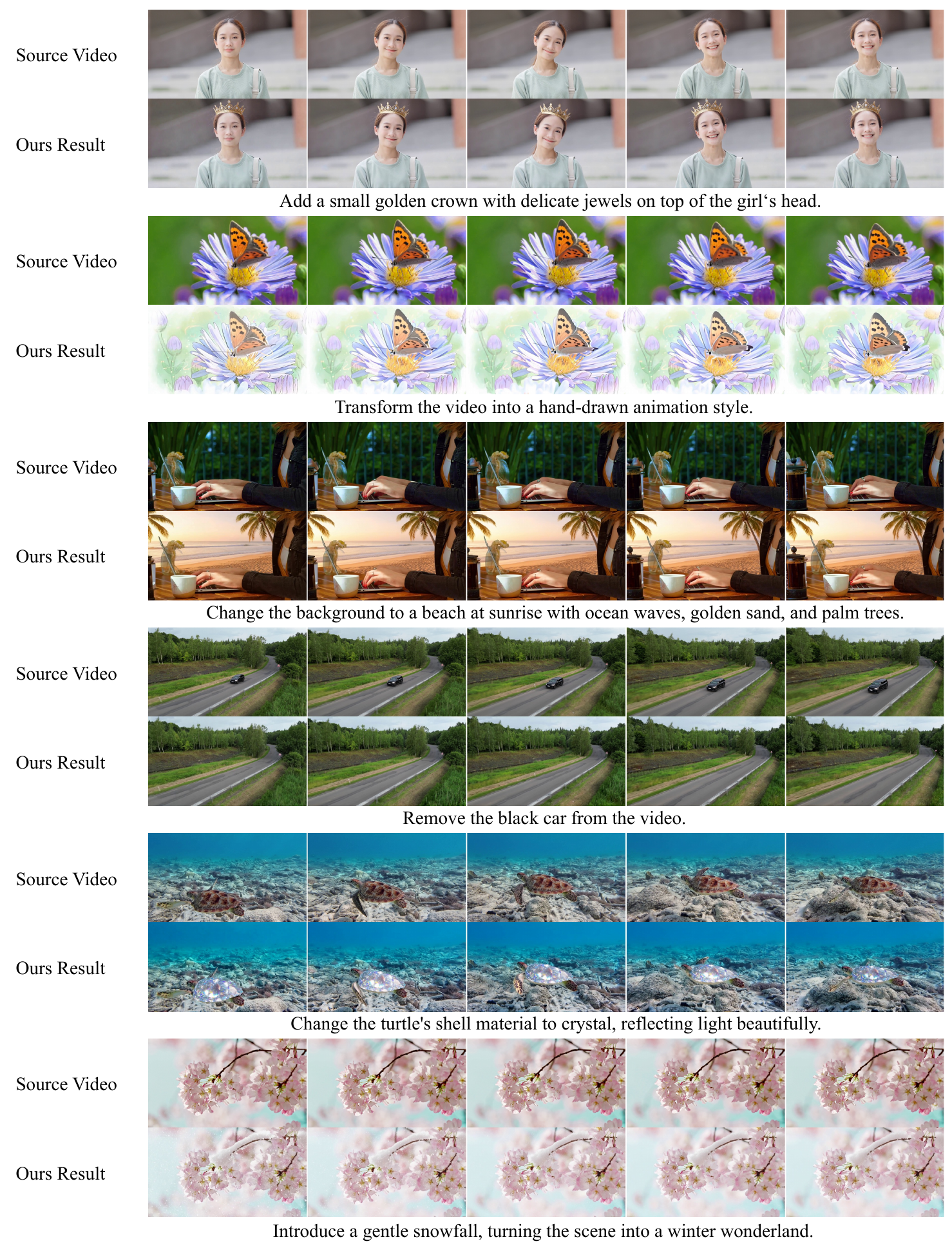}
  \vspace{-0.1in}
  \caption{More visual results in the landscape orientation on EditVerseBench.}
  \vspace{-0.2in}
  \label{fig:landscape_editverse}
\end{figure*}

\begin{figure*}[t]
  \centering
  \includegraphics[width=0.95\textwidth]{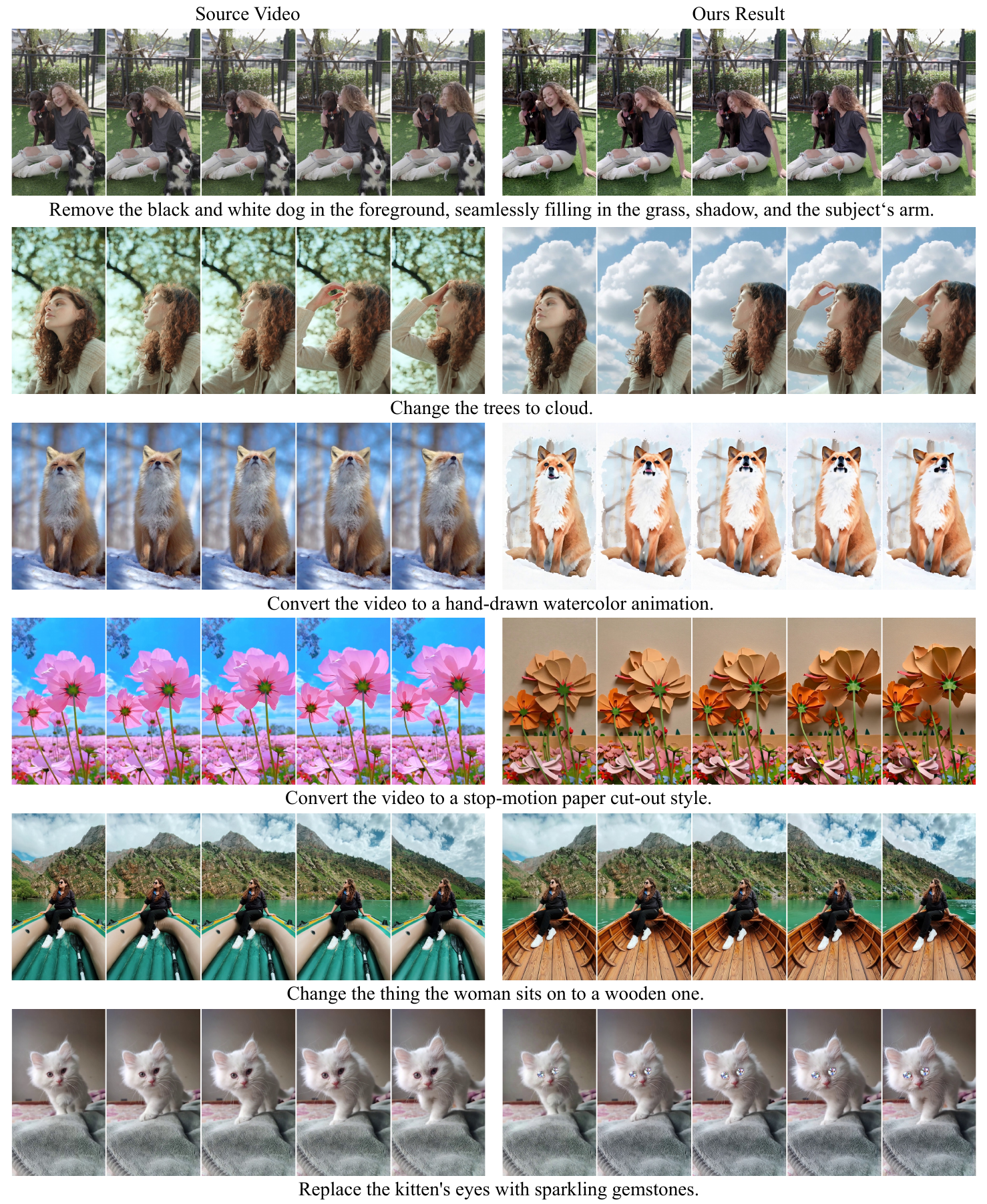}
  \vspace{-0.05in}
  \caption{More visual results in the portrait orientation on EditVerseBench.}
  \vspace{-0.2in}
  \label{fig:portrait_editverse}
\end{figure*}

As a complement to the visual examples in the main paper, we provide additional visualization results on EditVerseBench to offer a broader view of the editing results produced by our method. Among these results is a full-task visualization that shows all four main editing tasks—add, remove, change, and stylization—together with the corresponding source video, as shown in Fig.~\ref{fig:sup_fulltask_editversebench}. 
In addition, we include two orientation-specific visualizations: one for landscape orientation, as presented in Fig.~\ref{fig:landscape_editverse}, and one for portrait orientation, as illustrated in Fig.~\ref{fig:portrait_editverse}. Each visualization compares the edited videos with its corresponding source video and serves as a supplementary demonstration of our method's editing results under different video orientations.

\subsection{More Results on UNICBench}
\begin{figure*}[t]
  \centering
  \includegraphics[width=1\textwidth]{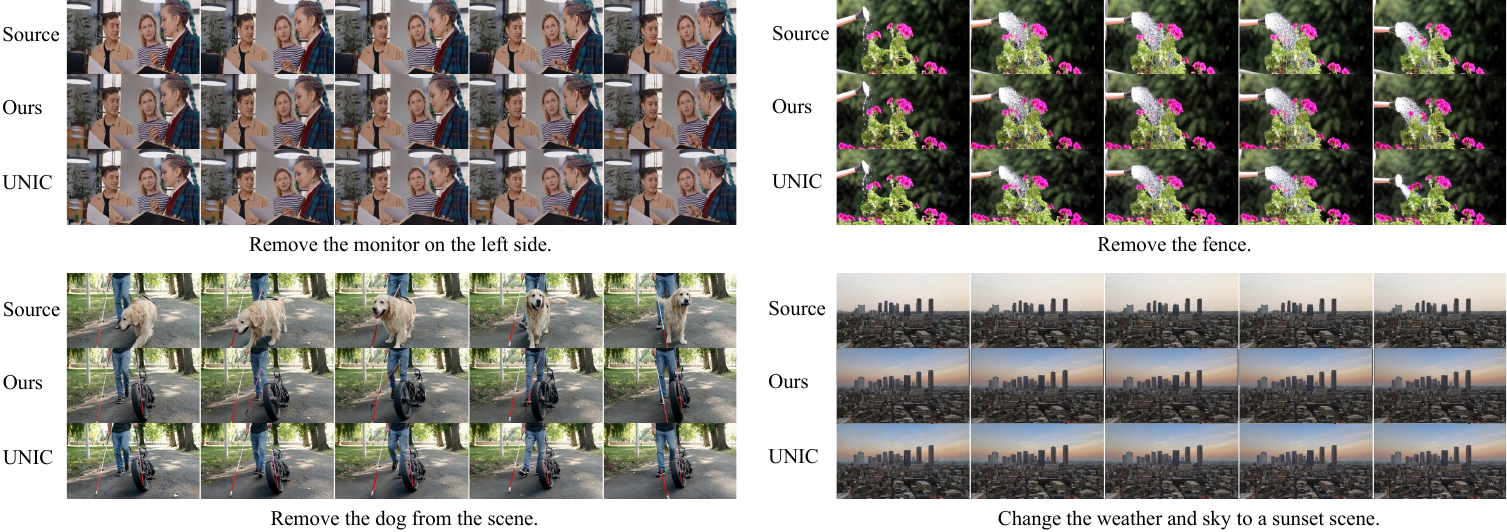}
  \vspace{-0.2in}
  \caption{Visualization results of our method and UNIC on UNICBench for FFP-based video editing.}
  \vspace{-0.2in}
  \label{fig:ffp_unic}
\end{figure*}

We provide additional visualization results on UNICBench to present the editing results of our method together with the source video and UNIC under the FFP-based video editing paradigm, as shown in Fig.~\ref{fig:ffp_unic}. This example offers a direct visual comparison of the editing results produced by our method and UNIC. 
We also include a mixed visualization that incorporates cases for which UNIC does not provide FFP-based video editing outputs, as presented in Fig.~\ref{fig:mix_unic}. In these cases, we present the instruction-based outputs from UNIC alongside our FFP-Based results to provide a broader visual reference across the different video editing types.


\end{document}